\documentclass[compsoc, conference, a4paper, 10pt, times]{IEEEtran}

\usepackage{cite}
\usepackage{amsmath,amssymb,amsfonts}
\usepackage{algorithmic}
\usepackage{graphicx}
\usepackage{textcomp}
\usepackage{xcolor}
\usepackage{booktabs}
\usepackage[hidelinks]{hyperref}
\usepackage{array}
\newcolumntype{P}[1]{>{\centering\arraybackslash}p{#1}}
\newcolumntype{M}[1]{>{\centering\arraybackslash}m{#1}}
\usepackage{paralist}
\usepackage[linesnumbered,ruled,vlined]{algorithm2e}
\usepackage{multirow}
\usepackage{subcaption}
\usepackage{enumitem}

\usepackage[flushleft]{threeparttable}
\usepackage{todonotes}

\begin{document}

\date{}

\title{Watermarking Graph Neural Networks based on Backdoor Attacks}

\author{\IEEEauthorblockN{Jing Xu}
\IEEEauthorblockA{\textit{Delft University of Technology} \\
Delft, The Netherlands \\
j.xu-8@tudelft.nl}
\and
\IEEEauthorblockN{Stefanos Koffas}
\IEEEauthorblockA{\textit{Delft University of Technology} \\
Delft, The Netherlands \\
s.koffas@tudelft.nl}
\and
\IEEEauthorblockN{Oguzhan Ersoy}
\IEEEauthorblockA{\textit{Radboud University} \\
Nijmegen, The Netherlands \\
oguzhan.ersoy@ru.nl}
\and
\IEEEauthorblockN{Stjepan Picek}
\IEEEauthorblockA{\textit{Radboud University} \\
Nijmegen, The Netherlands \\
stjepan.picek@ru.nl}
}
 
\maketitle


\begin{abstract}
Graph Neural Networks (GNNs) have achieved promising performance in various real-world applications. Building a powerful GNN model is not a trivial task, as it requires a large amount of training data, powerful computing resources, and human expertise in fine-tuning the model.  Moreover, with the development of adversarial attacks, e.g., model stealing attacks, GNNs raise challenges to model authentication. To avoid copyright infringement on GNNs, verifying the ownership of the GNN models is necessary.

This paper presents a watermarking framework for GNNs for both graph and node classification tasks. We 1) design two strategies to generate watermarked data for the graph classification task and one for the node classification task, 2) embed the watermark into the host model through training to obtain the watermarked GNN model, and 3) verify the ownership of the suspicious model in a black-box setting. The experiments show that our framework can verify the ownership of GNN models with a very high probability (up to $99\%$) for both tasks. Finally, we experimentally show that our watermarking approach is robust against a state-of-the-art model extraction technique and four state-of-the-art defenses against backdoor attacks.
\end{abstract}

\section{Introduction}
\label{sec:introduction}






Many real-world data can be modeled as graphs, e.g., social networks, gene interactions, and transport networks. 
Similar to the great success of deep learning algorithms on, e.g., image recognition~\cite{he2016deep, krizhevsky2012imagenet, simonyan2014very}, speech recognition~\cite{graves2013speech, hinton2012deep}, and natural language processing~\cite{goldberg2016primer}, deep graph models such as graph neural networks (GNNs)~\cite{kipf2017semi, velickovic2018graph, hamilton2017inductive} have also achieved promising performance in processing graph data. Such successful results can be attributed to their superior ability to incorporate information from neighboring nodes in the graph recursively~\cite{Wu2021}. 
Still, building and training a well-performed graph neural network is not a trivial task, as it usually requires a large amount of training data, effort in designing and fine-tuning a model, and powerful computing resources, making the trained model have a monetary value. For instance, the cost of training a machine learning model can be more than one million USD~\cite{strubell2019energy}.

As graph neural networks are more widely developed and used, their security also becomes a serious concern.
For instance, the adversary can steal the model through a model stealing attack.
Recent works have shown the high effectiveness of model stealing attacks on complex models even without knowledge of the victim's architecture or the training data distribution~\cite{tramer2016stealing, orekondy2019knockoff, papernot2017practical}, which leads to model copyright infringement. 
Moreover, if the model is intended to be released for commercial purposes, the stolen model would even lead to financial loss. Therefore, it is crucial to verify the ownership of a GNN model.

Digital watermarking is typically used to identify ownership of the copyright of media signals, e.g., audio, video, and image data~\cite{langelaar2000watermarking}. 
There are also works discussing embedding watermarks into DNN models to protect the IP of the models. For instance, Uchida et al.~\cite{uchida2017embedding} presented a framework to embed watermarks into the parameters of DNNs via the parameter regularizer during training leading to its white-box setting.  
To address the limitations of watermarking DNNs in the white-box setting, Adi et al. used random training instances and random labels to watermark a neural network in a black-box way~\cite{adi2018turning}. There, the authors based their approach on backdoor attacks.
Additionally, Zhang et al.~\cite{zhang2018protecting} extended the threat model to support black-box setting verification for DNN models. 
The general idea of watermarking a model in a black-box setting is to train the model using specific samples so that the model can memorize the watermark information and be verified when predicting on these samples.  

The discussed works focus on the image domain, not graph data. 
The watermark generation methods and injecting position differ between image data and graph data~\cite{zhao2021watermarking, xu2021explainability}. Specifically, as non-Euclidean data, the graph has rich structural information that can be used to generate the watermark. Moreover, in the image domain, the watermark injecting position can be defined, which is impossible in graph data as there is no position information one can exploit in a graph. 
Therefore, the discussed watermarking mechanism can only generate a watermark in image form but cannot be embedded into a graph. 

To the best of our knowledge, only one work considers watermarking graph data and GNNs. Zhao et al. presented a watermarking framework for GNNs by generating a random graph associated with features and labels as the watermark~\cite{zhao2021watermarking}. However, this work only studies the watermarking in GNNs for the node classification task while neglecting other relevant settings, e.g., the graph classification task.
Furthermore, the presented method only works for the GNN models trained through inductive learning.
In recent years, various transductive approaches have been widely applied and implemented in many domains due to their capability to adapt to different real-world applications, such as natural language processing, surveillance, graph reconstruction, and ECG classification~\cite{rossi2018inductive}. Therefore, it is possible that the owner's model is trained by transductive learning, and then the proposed method is not feasible as the graph structure of the training graph has been changed.

In this paper, we present a novel watermarking framework for GNNs suitable for both graph and node classification tasks as well as models trained by both inductive and transductive learning. 
Following the idea of~\cite{adi2018turning}, our watermarking method utilizes backdoor attacks. 
Backdoor attacks in GNNs aim to misclassify graph data embedded with a trigger.
In our work, instead of considering backdoor attacks in GNNs~\cite{weber2019anti,xi2021graph, zhang2021backdoor} for offensive purposes, we use them to protect the IP of the GNN models. More precisely, we use the backdoor triggers as digital watermarks to identify the ownership of a GNN model.
Our watermarking framework includes three phases: 
\begin{compactitem}
    \item \textbf{Watermarked data generation}. We designed two strategies to generate watermarked data for the graph classification task and one for the node classification task. 
    \item \textbf{Watermark embedding}. We train the host model with the watermarked data. The intuition here is to explore the memorization capabilities of GNNs to automatically learn the trigger pattern of the watermarked data. 
    \item \textbf{Ownership verification}. Once the watermark is embedded into the model, we can verify the ownership of remote suspicious models by sending watermarked data generated in the first phase. Only the models protected by the watermarks are assumed to output matched predictions. To address the limitations of~\cite{zhao2021watermarking}, we specifically use the feature trigger as the watermark pattern by modifying the feature information of the graph instead of changing the graph's structure.
\end{compactitem}

We evaluate our watermarking framework with five benchmark datasets: two for the node classification task and three for the graph classification task. The results show that our watermarking framework can verify the ownership of suspicious models with high probability. At the same time, the performance of the watermarked GNN on its original task can be preserved.
Our main contributions can be summarized as follows:
\begin{compactitem}
    \item We propose a watermarking framework to verify the ownership of GNN models for both the node classification task and the graph classification task. It is the first watermarking framework for GNNs on the graph classification task. 
    \item We use hypothesis testing in our watermarking mechanism to provide statistical analysis for the model ownership verification results.
    \item We propose two watermark generation mechanisms to generate watermarked data for the graph classification task. One strategy is based on classical backdoor attacks, and the other is based on embedding the watermark into random graphs, which experimentally shows superior performance.
    \item For the node classification task, we propose a training-agnostic framework that is also applicable to a model trained by transductive learning. Specifically, we only modify the feature information of the graph in the watermarked data generation phase. 
    \item We evaluate the proposed watermarking framework with several benchmark datasets and popular GNN models. 
    Experimental results show that the proposed method achieves excellent performance, i.e., up to $99\%$ accuracy, in IP protection of the models while having a negligible impact on the original task (less than $1\%$ clean accuracy drop). 
    \item We investigate the robustness of our method against a model extraction attack and four defenses against backdoor attacks. Experimental results show our watermarked model is robust against these mechanisms. 
\end{compactitem}

\section{Background}
\label{sec:background}


\subsection{Graph Neural Networks (GNNs)}

GNNs take a graph $G$ as an input (including its structure information and node features) and learn a representation vector (embedding) for each node $v \in G$, $z_v$, or the entire graph, $z_G$. Modern GNNs follow a neighborhood aggregation strategy, where one iteratively updates the representation of a node by aggregating representations of its neighbors. After $k$ iterations of aggregation, a node's representation captures both structure and feature information within its $k$-hop network neighborhood. Formally, the $k$-th layer of a GNN is (e.g., GCN~\cite{kipf2017semi}, GraphSAGE~\cite{hamilton2017inductive}, and GAT~\cite{velickovic2018graph}):
\begin{equation}
    Z^{(k)} = AGGREGATE(A, Z^{(k-1)};\theta^{(k)}).
    \label{eqn:2.1-1}
\end{equation}
Here, $Z^{(k)}$ represents the node embeddings in the matrix form computed after the $k$-th iteration, and the $AGGREGATE$ function depends on the adjacency matrix $A$, the trainable parameters $\theta^{(k)}$, and the previous node embeddings $Z^{(k-1)}$. Finally, $Z^{(0)}$ is initialized as $G$'s node features.

For the node classification task, the node representation $Z^{(k)}$ of the final iteration is used for prediction, while for the graph classification task, the $READOUT$ function pools the node embeddings from the final iteration $K$: $z_G = READOUT(Z^{(K)}).$
$READOUT$ can be a simple permutation invariant function or a more sophisticated graph-level pooling function~\cite{ying2018hierarchical, zhang2018end}.

The goal of the node classification task is that, given a single graph with partial nodes being labeled and others remaining unlabeled, GNNs can learn a robust model that effectively identifies the class labels for the unlabeled nodes~\cite{kipf2017semi}. In a node classification task, there are two types of training settings - inductive and transductive. In an inductive setting, the unlabeled nodes are not seen during training, while in a transductive setting, the test nodes (but not their labels) are also observed during the training process. The node classification task is used in many security applications, e.g., anomaly detection in the Bitcoin transaction network or terrorist detection in a social network~\cite{chaudhary2019anomaly}.
The graph classification task aims to predict the class label(s) for an entire graph~\cite{zhang2018end}. One practical application of the graph classification task is to detect whether a molecule is a mutagen or not~\cite{kriege2012subgraph}.
Therefore, it is relevant to consider both classification tasks when considering real-world applications.

\subsection{Backdoor Attacks in GNNs} 

Deep Neural Networks (DNNs) are vulnerable to backdoor attacks~\cite{liu2017trojaning, li2020invisible}. Specifically, a backdoored neural network classifier produces attacker-desired behaviors when a trigger is injected into a test example. Several studies showed that GNNs are also vulnerable to backdoor attacks.
Similar to the idea of backdoor attack in DNNs, the backdoor attack in GNNs is implemented by poisoning the training data with a trigger, which can be a subgraph with/without features~\cite{zhang2021backdoor, xi2021graph} or a subset of node features~\cite{xu2021explainability}. After training the GNN model with the trigger-embedded data, the backdoored GNN would predict the test example injected with a trigger as the pre-defined target label.


\subsection{Digital Watermarking in Neural Networks}

Digital watermarking is a technique that embeds certain watermarks in carrier multimedia data such as audio, video, or images to protect their copyright~\cite{langelaar2000watermarking}. The information to be embedded in a signal is called a digital watermark. The signal where the watermark is embedded is called the host signal. A digital watermarking system is usually divided into two steps: embedding and verification. The typical digital watermarking life cycle is presented in Appendix~\ref{appendix:digital_watermarking_life_cycle}.  

At first, the goal of digital watermarking was to protect the copyright of multimedia data by embedding watermarks into the multimedia data. More recently, with the development of deep neural networks, new watermarking methods were designed to protect the DNN models by embedding watermarks into DNN models~\cite{uchida2017embedding, chen2019deepmarks, adi2018turning}. The idea of watermarking neural networks is similar to traditional digital watermarking in multimedia data. 
To implement digital watermarking in neural networks, we can assume the multimedia data that we want to protect is the model, and the embedding and verification steps of the watermarking correspond to the training and inference phase of the protected model. 
A neural network watermarking model should satisfy the following requirements~\cite{DBLP:journals/ijon/LiWB21,DBLP:journals/fdata/Boenisch21}: robustness, fidelity, capacity, integrity, generality, efficiency, and secrecy. 
The details about the requirements are given in Section~\ref{sec:watermark_requirements}.




\section{GNN Watermarking}
\label{sec:gnn_watermarking}

\subsection{General Idea}

\begin{figure}
\centering
\includegraphics[width=0.48\textwidth, page=2]{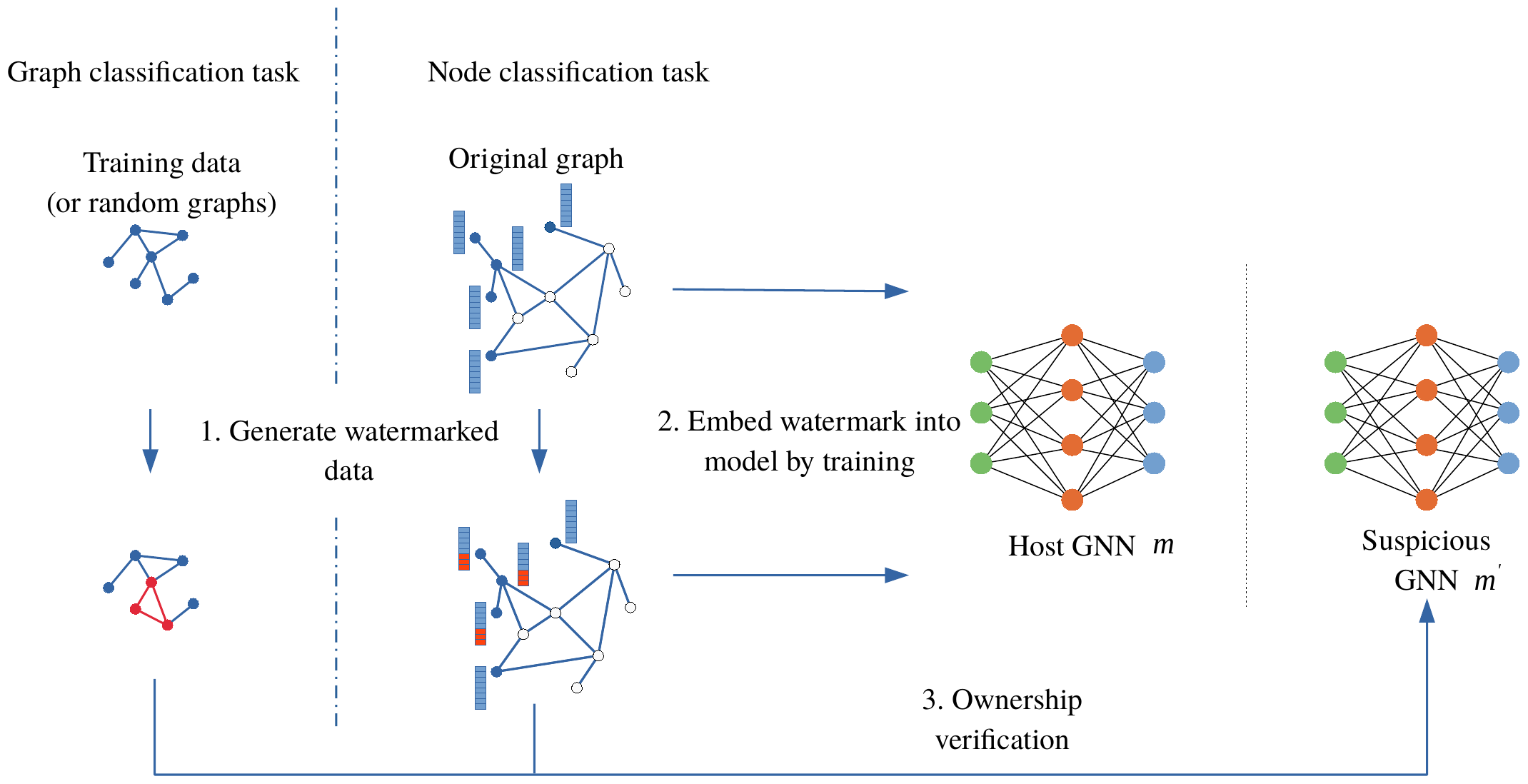}
\caption{\label{fig:gnn_watermark_framework}GNN watermarking framework.}
\end{figure}

This section discusses how to apply the backdoor attack to watermark GNN models. More precisely, we propose a framework to generate watermarked data, embed a watermark into GNNs, and verify the ownership of GNNs by extracting a watermark from them. The framework's purpose is to protect the IP of the graph neural networks by verifying the ownership of suspicious GNNs with an embedded watermark. The framework first generates watermarked data and trains the host GNNs with the watermarked data. Through training, the GNNs automatically learn and memorize the connection between the watermark pattern and the target label. As a result, only the model protected with our watermark can output predefined predictions, i.e., the assigned target label, when the watermark pattern is observed in the queries sent to the suspicious model. Figure~\ref{fig:gnn_watermark_framework} illustrates the workflow of our GNN watermarking framework. 


\subsection{Threat Model}
\label{threat_model}

In our threat model, we model two parties, a model owner, who owns a graph neural network model $m$ for a certain task $t$, and a suspect, who sets up a similar service $t^{'}$ from the model $m^{'}$, where two services have an alike purpose $t\approx t^{'}$. In practice, there are multiple ways for a suspect to get the model $m$. For example, it could be an insider attack from the owner's organization that leaks the model, or it could be stolen by various model stealing attacks, e.g.,~\cite{tramer2016stealing,batina2019csi}. Although the exact mechanism of how a suspect obtains the model $m$ is out of the scope of this paper, we still evaluate our watermarking method against a model extraction attack in Section~\ref{sec:robustness}. 

Our goal is to help the model owner protect his/her model $m$, which is an intellectual property with a concrete value. Intuitively, if model $m$ is equivalent to $m^{'}$, we can confirm that the suspect is a plagiarizer and $m^{'}$ is a plagiarized model of $m$.\footnote{The chances that various entities created the same model independently are small, especially when considering real-world applications.} We define the owner's model $m$ as the host model and the model $m^{'}$, which is likely to be stolen from $m$ as the suspicious model. 
In this work, we assume that as an IP protector, we can only query the suspicious model $m^{'}$ in a black-box manner.
As the model owners, we have full access to the model $m$, including its architecture, training data, and the training process.

\subsection{Watermarked Data Generation}
\label{watermark_generation}

\textbf{Graph classification.}
Since most graph classification tasks are implemented by utilizing GNNs to learn the network structure, we focus on utilizing the subgraph-based backdoor attacks~\cite{zhang2021backdoor,xi2021graph} to verify the ownership of GNNs for the graph classification task. Here, we investigate two watermarking data generation strategies for GNNs on the graph classification task.

\begin{enumerate}[label=(\roman*)]
\item \textbf{Embedding watermark into original training data}. Specifically, we select a subset of samples in training data and embed a generated random watermark (i.e., a random graph) into it. A random graph including $n$ nodes and $e$ edges is generally generated by starting with $n$ isolated nodes and inserting $e$ edges between nodes at random. A classical method to generate random graphs is called Erdos-Renyi (ER) random graph model~\cite{Gilbert1959} in which each possible edge occurs independently with a probability $p \in (0, 1)$. This paper utilizes ER random graphs as the watermark graph for watermarking GNNs on the graph classification task.
Once the watermark graph is generated, we embed it into each graph of the selected subset of training data by randomly choosing $n$ nodes in the graph and changing their connections to be the same as the watermark graph. Since we only change the graph's structure, we do not modify the node's features. The watermark information is also carried by the label of the watermark embedded graphs. We assume the value of the label of the watermarked graphs is in the range $\left [0, C \right )$, where $C$ is the number of classes, and the label for watermarked graphs can be determined in advance. We emphasize that the labels of the sampled training data are different from the label of the watermark embedded graphs. In this way, three parameters - $r$ (proportion of training data selected to be injected with the watermark graph), $n$ (number of nodes in the watermark graph), and $p$ (the probability of the edge existence in the watermark graph) would have a significant impact on the watermark generation, which will then affect the watermark embedding and verification later.

\item \textbf{Embedding watermark into generated random graphs}. In addition to embedding a watermark graph into the original training data, we propose first generating random graphs and then embedding a generated watermark into these random graphs. The intuition here is that by embedding a watermark graph into the original training data, as discussed in the previous paragraph, the watermark will have some side effects on the original functionality of our watermarked graph neural networks. We design this strategy to decrease the impact of watermarking in the original task.
First, we generate a number of random graphs with the ER method, where we define that number as a specific proportion ($r$) of the training data. The number of nodes and edge existence probability are the same as the average number of nodes and edges of the training data. Then, we generate the watermark graph in the same way described in the previous approach and embed the watermark graph into the random graphs generated in the first step. We use the node degree as the node feature for the generated random graphs. We also assign the label for the watermark embedded graphs in advance, similar to the first strategy. 
There are also three parameters - $r$, $n$, and $p$ in this strategy.
\end{enumerate}

The detailed comparison and analysis of these parameters in the above two strategies are shown in Section~\ref{sec:experiments}.

\textbf{Node classification.}
We apply the backdoor attack as proposed in~\cite{xu2021explainability} to implement watermarking GNNs for the node classification task, which can be applied in not only inductive learning but also transductive learning-based models. Specifically, we randomly select a proportion $r$ of the total number of nodes in the graph as the watermark carrier nodes and change their subset node features\footnote{Here, the number of node features whose values are changed is defined as watermark length $l$.} into a predefined fixed value\footnote{The fixed value is uniformly selected between $0$ and $1$.} to generate the watermarked data. Given an arbitrary node in the graph, by changing the value of a subset of its features as a feature trigger and assigning a target label to it, the host model $m$ aims to learn and memorize the watermark pattern.

\subsection{Watermark Embedding}
\textbf{Graph classification.} Once the watermarked data are generated, the next step in the framework is to embed the watermark into the host GNN model $m$. Here, we explore the intrinsic learning capability of graph neural networks to embed the watermark.
We first train a clean model $m_c$ based on the original training data $D_{train}$ and then continue training the model using the watermarked data. 
The detailed GNN watermark embedding process is shown in Algorithm~\ref{alg:watermark_embedding}. The inputs are the pre-trained clean model $m_c$, original training data $D_{train}$ and target label for the watermarked data, and the outputs are the watermarked GNN model $m_w$ and watermarked data $D_{wm}$. The model owner defines the target label of the watermarked data. In the main function, we sample data $D_{tmp}$ from the original training data uniformly at random.
The data we sample has a label that is different from the target label (Line 3 in Algorithm~\ref{alg:watermark_embedding}) so that we can avoid the influence of the original label in the ownership verification phase. 
For the second watermarking strategy for the graph classification task, which is based on generating random graphs as watermark carrier data, we utilize the $ER$ method to generate random graphs with an average number of nodes and edges of the training data and proportion $r$ (Line 5 in Algorithm~\ref{alg:watermark_embedding}). Then, for each data in $D_{tmp}$, we add the generated watermark to $x$ and relabel it with $y_{t}$ (Lines 7-11 in Algorithm~\ref{alg:watermark_embedding}). Therefore, we obtain the watermarked data $D_{wm}$, which is later used in the verification process. 
We train the pre-trained clean GNN model with both the sampled original data $D_{tmp}$ and $D_{wm}$ (or just $D_{wm}$ for the second strategy).\footnote{We use both the sampled original data and watermarked data for the first strategy to decrease the impact of watermarking on the model's original main task.}
We assume that the GNN model will learn the watermark pattern during the training process and, thus, be protected against the model stealing attacks.


\begin{algorithm}
\small
\SetAlgoLined
\caption{Watermark embedding for graph classification task} \label{alg:watermark_embedding}
\SetKwInput{KwInput}{Input}
\SetKwInput{KwOutput}{Output}
\DontPrintSemicolon
    \KwInput{
    Pre-trained clean model $m_c$, Training set $D_{train} = \left \{ x_i, y_i \right \}_{i=1}^{S}$, Target label $y_t \in \left [0, C\right )$\\}
    \KwOutput{
    Watermarked GNN model $m_w$, Watermark data $D_{wm}$\\
    }
    
    \SetKwFunction{FMain}{WATERMARK\_EMBEDDING()}
    
    \SetKwProg{Fn}{Function}{:}{}
    \Fn{\FMain}{
        $D_{wm} \leftarrow \emptyset $\;
        $D_{tmp} \leftarrow sample(D_{train}, r, y \neq y_t)$ // strategy 1\;
        // or\;
        $D_{tmp} \leftarrow GRAPH\_GENERATE(n_{avg}, p_{avg}, r))$ // strategy 2\;
        \ForEach{$d \in D_{tmp}$}
        {
            $x_{wm} = ADD\_WATERMARK(d[x], watermark)$\;
            $y_{wm} = y_t$\;
            $D_{wm} = D_{wm} \cup \left\{x_{wm}, y_{wm} \right\}$\;
        }
    }
    \textbf{End Function}\;
    $m_w = Train(m_c, D_{wm}, D_{tmp})$\;
    (or $m_w = Train(m_c, D_{wm})$)\;
    \textbf{return} $m_w, D_{wm}$
\end{algorithm}

The watermarking embedding process for the node classification is the same as the graph classification task.

\subsection{Ownership Verification}
\label{subsec:ownership_verification}
After training our model with watermarked data, if adversaries steal and further fine-tune the watermarked model, they will likely set up an online service to provide the AI service of the stolen model.
Then, it is difficult to access the architecture and parameters of the suspicious model directly.
However, as we have explained in Section~\ref{threat_model}, to verify the ownership of the suspicious model $m^{'}$ in a black-box manner, we can send $D_{wm}$, which is returned in the previous watermark embedding process to the suspicious model. 
If for part of samples in $D_{wm}$, the suspicious model outputs the target label $y_{wm}$, we can assume that $m^{'}$ is stolen (developed) from our watermarked model $m_w$. 
However, the premise of this assumption is that our watermarked model has statistically different behavior from the clean model, leading to different watermark accuracy between these two models. 
To provide a statistical guarantee with the model ownership verification results, we can adopt statistical testing with the ability to estimate the level of confidence to determine whether the watermark accuracy of our watermarked model is significantly different from the clean model. 
We define the null hypothesis $\mathcal{H}_0$ as follows: 
\[ \mathcal{H}_0: P_r(m_w(x_{wm}) = y_{wm}) \cong P_r(m_c(x_{wm}) = y_{wm}) , \]
where $P_r(m_w(x_{wm}) = y_{wm})$ represents the watermark success probability of the watermarked model, and $P_r(m_c(x_{wm}) = y_{wm})$ represents the watermark success probability of a clean model.

The null hypothesis $\mathcal{H}_0$ states that the watermark success probability of the watermarked model is equal or approximate to the clean model, i.e., there are no significant differences between the watermark accuracy of the watermarked model and a clean model. On the contrary, the alternative hypothesis $\mathcal{H}_1$ states that the watermarked model has significantly different watermark accuracy from the clean model, which verifies the effectiveness of our watermarking mechanism. If we can reject the null hypothesis $\mathcal{H}_0$ with statistical guarantees, we can claim that our watermarking method successfully verifies the ownership of the suspicious models.

Through querying a series of watermarked and clean models with $q$ watermarked samples, we can obtain their prediction results for each watermarked model and clean model, which can be used to calculate the watermark accuracy, denoted as $\alpha_k$ and $\beta_k$, respectively: 

\begin{align}
    \alpha_k = \frac{\sum_{i=1}^{q}\mathbb{I}(y_i^{w_k}=y_{wm})}{q}, k\in[1, K] \\
    \beta_k = \frac{\sum_{i=1}^{q}\mathbb{I}(y_i^{c_k}=y_{wm})}{q}, k\in[1, K]
\end{align}
where $K$ is the number of watermarked and clean models. We set $K$ to $10$ in our experiments.

The value of watermark accuracy can be considered as an estimation of the watermark success probability. 
We apply the Welch's t-test~\cite{welch1947generalization}\footnote{We use the Welch's t-test since the watermark accuracy of clean and watermarked models can be treated as normal distributions according to a Shapiro-Wilk Test~\cite{shapiro1965analysis}, and they may have different variances.} to test the hypothesis.
According to the watermark accuracy of $K$ watermarked models and clean models, denoted as $\left\{\alpha_1,\cdots \alpha_n \right\}$ and $\left\{\beta_1,\cdots \beta_n \right\}$ respectively, we can calculate the $t$ statistic as follows:
\begin{equation}
\label{equ:t_statistic}
    t=\frac{\bar{\alpha}-\bar{\beta}}{\sqrt{\frac{s_{\alpha}^2}{n}+\frac{s_{\beta}^2}{n}}}, 
\end{equation}
where $s_{\alpha}^2$ and $s_{\beta}^2$ are the unbiased estimators of the population variance. 

The degrees of freedom $\nu$ associated with variance estimate is approximated using the Welch-Satterthwaite equation~\cite{satterthwaite1946approximate, welch1947generalization}:
\begin{equation}
\label{equ:nu}
   \nu \approx \frac{(n-1)(s_{\alpha}^2+s_{\beta}^2)^2}{s_{\alpha}^4+s_{\beta}^4}.
\end{equation}

According to the theoretical analysis above, we can formally state under what conditions the model owner can reject the null hypothesis $\mathcal{H}_0$ at the significance level $1-\tau$ (i.e., with $\tau$ confidence) with watermark accuracy of watermarked and clean models. 
Specifically, we take the watermark accuracy results of NCI1 and DiffPool as an example, as shown in Table~\ref{Table:t_test}. 
According to a Shapiro-Wilk Test~\cite{shapiro1965analysis}, the $p$-values~\cite{wasserstein2016asa} of these two populations, i.e., watermark accuracy of clean and watermarked models, are $0.77$ and $0.16$, respectively. 
Given a significance level of $0.05$, these $p$-values indicate these two populations can be assumed to be normally distributed, and a Welch's t-test is applicable. 
With significance level $1-\tau=0.05$, and the degree of freedom calculated with Eq.~\eqref{equ:nu}, i.e., $\nu\approx16$, the $t$ critical value $t_\tau$ is $2.120$. Based on Eq.~\eqref{equ:t_statistic}, we calculate $t$ statistic $t=45.82$, which is significantly larger than $t_\tau=2.120$. Thus, we can reject the null hypothesis $\mathcal{H}_0$ at the significance level $0.05$ for the NCI1 dataset on the DiffPool model in our work. The watermark accuracy of the watermarked models and clean models of other datasets and models are presented in Appendix~\ref{appendix:t_test}. For each setting, we can reject the null hypothesis $\mathcal{H}_0$ at the significance level $0.05$, which provides a statistical guarantee for our watermarking method. 
Based on the statistical analysis above, we can also calculate a threshold for each dataset and model to ensure a low false positive rate (FPR) and false negative rate (FNR), i.e., less than $0.0001$, of our watermarking method. More details are presented in Section~\ref{sec:experiments}. 

\begin{table*}[!ht]
 \caption{Watermark accuracy of the watermarked and clean models on NCI1 with DiffPool ($n=10$).}
\begin{center}
\begin{tabular}{ccccccccccc}
 \hline
 {Models} & \multicolumn{10}{c}{Watermark Accuracy (\%)} \\
 \hline
 Clean & $0.70$ & $6.49$ & $4.27$ & $9.77$ & $14.85$ & $7.43$ & $1.59$ & $5.71$ & $14.09$ & $10.94$ \\
 \hline
 Watermarked & $94.98$ & $94.88$ & $99.32$ & $92.67$ & $92.11$ & $99.73$ & $99.58$ & $89.52$ & $99.78$ & $97.50$ \\
 \hline
\end{tabular}
\label{Table:t_test}
\end{center}
\end{table*}

\section{Evaluation}
\label{sec:experiments}

This section presents experimental results on different datasets and GNN models, a comparison with state-of-the-art, and a discussion on watermarking requirements. We run the experiments on a remote server with one NVIDIA 1080 Ti GPU with 32GB RAM. We use the PyTorch framework for the experiments, and each experiment is repeated ten times to obtain the average result.


\textbf{Dataset.} For the graph classification task, we use three publicly available real-world graph datasets, one chemical dataset, and two social datasets: (i) NCI1~\cite{morris2020tudataset} - a subset of the dataset consisting of chemical compounds screened for activity against non-small cell lung cancer. (ii) COLLAB~\cite{yanardag2015deep} - a scientific collaboration dataset, derived from three public collaboration datasets. (iii) REDDIT-BINARY~\cite{yanardag2015deep} - a dataset consisting of graphs corresponding to online discussions on Reddit.

For the node classification task, we also use two real-world datasets: Cora~\cite{sen2008collective} and CiteSeer~\cite{sen2008collective}. These two datasets are citation networks in which each publication is described by a binary-valued word vector indicating the absence/presence of the corresponding word in the collection of $1,433$ and $3,703$ unique words, respectively. Table~\ref{Table:data_statistics} shows the statistics of all considered datasets.

\begin{table*}
 \caption{Datasets statistics.}
\begin{center}
\begin{tabular}{cccccc}
 \hline
 Datasets & \# Graphs & Avg. \# nodes & Avg. \# edges & Classes & Class Distribution\\
 \hline
 \hline
 NCI1 & $4,110$ & $29.87$ & $32.30$ & $2$ & $2,053[0], 2,057[1]$ \\
 \hline
 COLLAB & $5,000$ & $74.49$ & $2,457.78$ & $3$ & $2,600[0], 775[1], 1,625[2]$ \\
 \hline
 REDDIT-BINARY & $2,000$ & $429.63$ & $497.75$ & $2$ & $1,000[0], 1,000[1]$ \\
 \hline
 \hline
 \multirow{2}{*}{Cora} & \multirow{2}{*}{$1$} & \multirow{2}{*}{$2,708$} & \multirow{2}{*}{$5,429$} & \multirow{2}{*}{$7$} & $351[0], 217[1], 418[2], 818[3],$ \\
 & & & & & $426[4], 298[5], 180[6]$ \\
 \hline
 \multirow{2}{*}{CiteSeer} & \multirow{2}{*}{$1$} & \multirow{2}{*}{$3,327$} & \multirow{2}{*}{$4,608$} & \multirow{2}{*}{$6$} & $264[0], 590[1], 668[2],$ \\
 & & & & & $701[3], 596[4], 508[5]$ \\
 \hline
\end{tabular}
\label{Table:data_statistics}
\end{center}
\end{table*}

\textbf{Dataset splits and parameter setting.} For each graph classification dataset, we sample $2/3$ as the training data and the rest as the test data. 
We set the watermark graph size $n$ as $\gamma$ fraction of the graph dataset's average number of nodes. We then sample or generate an $r$ fraction of the training data (with an un-target label) to embed the generated watermark. We explore the impact of these parameters in Section~\ref{results}.
For each node classification dataset, we use $20\%$ of total nodes as the training data. We set the size of the feature watermark to $l$ and then sample $r$ fraction of the training data to embed the generated feature watermark. The comparison of watermarking performance under different variants is shown in Section~\ref{results}.

\textbf{Models.} In our experiments, we use three the state-of-the-art GNN models for the graph classification task: DiffPool~\cite{ying2018hierarchical}, GIN~\cite{xu2018powerful}, and GraphSAGE~\cite{hamilton2017inductive}.
For the node classification task, we use GCN~\cite{kipf2017semi}, GAT~\cite{velickovic2018graph}, and GraphSAGE~\cite{hamilton2017inductive} as the host models.

\textbf{Metrics.} The main purpose of our watermarking framework is to verify the ownership of the suspicious GNN model successfully. 
According to the statistical analysis in Section~\ref{subsec:ownership_verification}, we can guarantee that our watermarking method can successfully verify the ownership of the suspicious models. Specifically, based on the watermark accuracy distribution of our watermarked models and clean models, we can calculate a threshold of watermark accuracy for each dataset and model to ensure a low FPR and FNR, i.e., less than $0.0001$, as shown in Tables~\ref{watermark_acc_threshold_graph} and~\ref{watermark_acc_threshold_node} for graph classification task and node classification task, respectively. 
In Table~\ref{watermark_acc_threshold_graph}, $D_{wm}^{t}$ is the watermarked data generated by embedding a watermark into sampled training data, while $D_{wm}^{r}$ is the watermarked data generated by embedding a watermark into the generated random graphs. 
As we can see from Table~\ref{watermark_acc_threshold_graph}, the watermark accuracy threshold of the second strategy is obviously higher than the first strategy. It can be explained that, in the first strategy, the clean model will likely classify the watermarked data into the original label since it is generated based on the training data, and the clean model does not learn the watermark pattern. 
However, in the second strategy, the watermarked data is generated based on random graphs, and the clean model is likely to classify it uniformly at random. Thus, the watermark accuracy of the clean models in the first strategy is nearly $0\%$ and that in the second strategy is around $1/C$ ($C$ is the number of classes). As a result, the watermark accuracy of the watermarked models in the second strategy should be higher than the first strategy to ensure the distinguishable difference between the watermarked models and the clean models. 
Once the watermark accuracy threshold is defined, we send queries of generated watermarked data to the suspicious model $m^{'}$. If the watermark accuracy of the suspicious model is over the corresponding threshold, we can reach the conclusion that the suspicious model is stolen or developed from the host model. 

\begin{table}
 \caption{Watermark accuracy threshold for each dataset and model on the graph classification task.}
\begin{center}
\begin{tabular}{cccc}
 \hline
 \multirow{2}{*}{Dataset} & \multicolumn{3}{c}{Watermark Acc. Threshold (\%) ($D_{wm}^{t}$ $|$ $D_{wm}^{r}$)}\\
 \cline{2-4}
 & {DiffPool} & {GIN} & {GraphSAGE} \\
 \hline
 \hline
 {NCI1} & {$53.5 | 72.0$} & {$44.0 | 76.5$} & $45.50 | 75.5$ \\
 \hline
 {COLLAB} & {$47.0 | 63.0$} & $44.5 | 62.5$ & $50.5 | 65.5$ \\
 \hline
 {REDDIT-} & \multirow{2}{*}{$49.5 | 71.0$} & \multirow{2}{*}{$46.5 | 68.0$} & \multirow{2}{*}{$48.5 | 76.0$} \\
 {BINARY} & & & \\
 \hline
\end{tabular}
\label{watermark_acc_threshold_graph}
\end{center}
\end{table}

\begin{table}
 \caption{Watermark accuracy threshold for each dataset and model on the node classification task.}
\begin{center}
\begin{tabular}{cccc}
 \hline
 \multirow{2}{*}{Dataset} & \multicolumn{3}{c}{Watermark Accuracy Threshold (\%)}\\
 \cline{2-4}
 & {GCN} & {GAT} & {GraphSAGE} \\
 \hline
 \hline
 {Cora} & $50.0$ & $51.0$ & $48.0$ \\
 \hline
 {CiteSeer} & $53.0$ & $48.0$ & $49.5$ \\
 \hline
\end{tabular}
\label{watermark_acc_threshold_node}
\end{center}
\end{table}

In addition to a good performance on the ownership verification task, a well-designed watermarking method should have only slight side effects on the host model's original task. We here also check whether our watermarking framework reduces the performance of the watermarked GNN model on its original task. Specifically, we compare watermarked and clean models' accuracy on the normal test data. 

\subsection{Experimental Results}
\label{results}

\begin{figure}[!ht]
     \centering
     \begin{subfigure}{0.48\textwidth}
         \centering
         \includegraphics[width=1\textwidth]{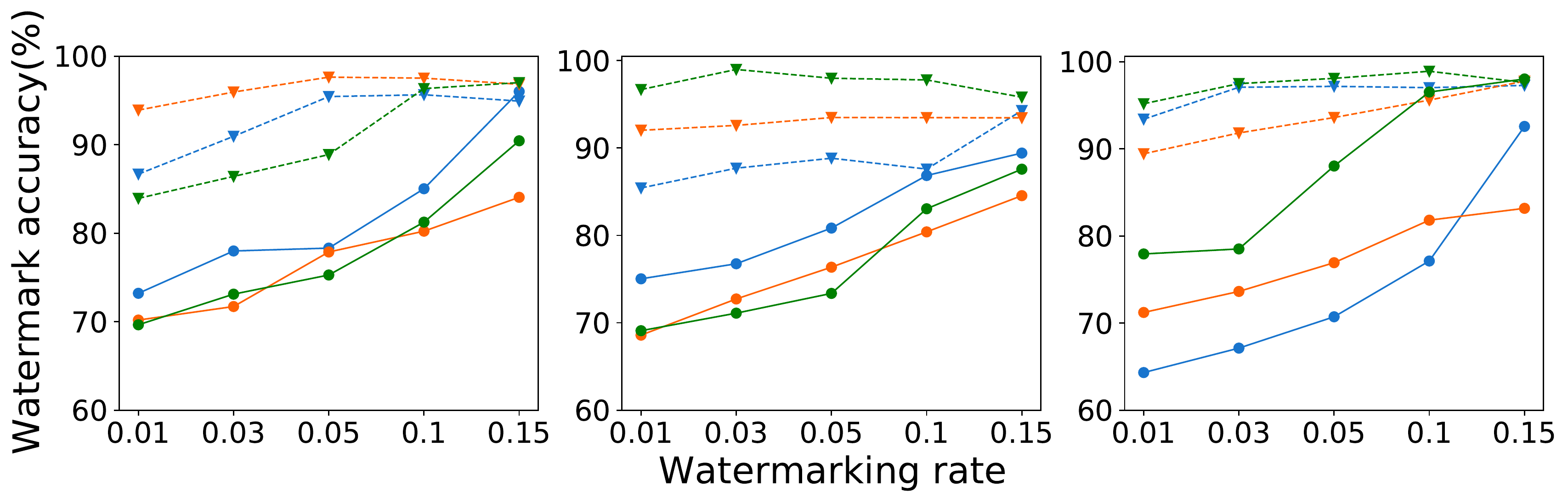}
         \caption{with different watermarking rate $r$ ($\gamma=0.2, p=1.0$)}
         \label{fig:asr_graph_classification-a}
     \end{subfigure}
     \newline
     \begin{subfigure}{0.48\textwidth}
         \centering
         \includegraphics[width=\textwidth]{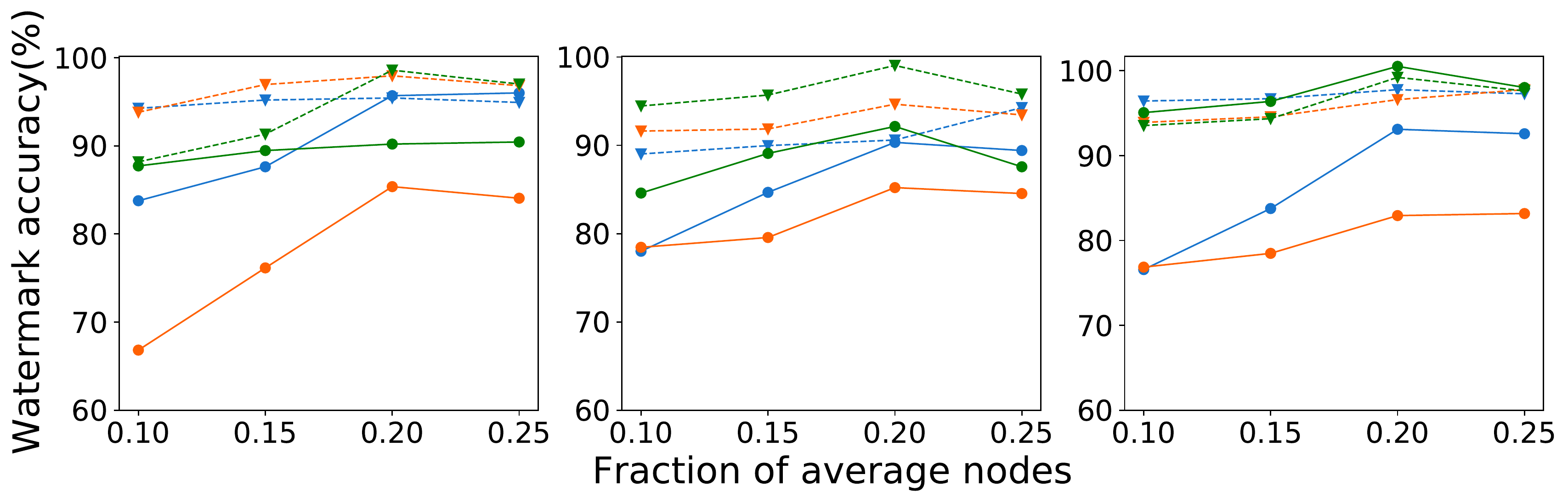}
         \caption{with different fraction of average nodes $\gamma$ ($r=0.15, p=1.0$)}
         \label{fig:asr_graph_classification-b}
     \end{subfigure}
     \newline
     \begin{subfigure}{0.48\textwidth}
         \centering
         \includegraphics[width=\textwidth]{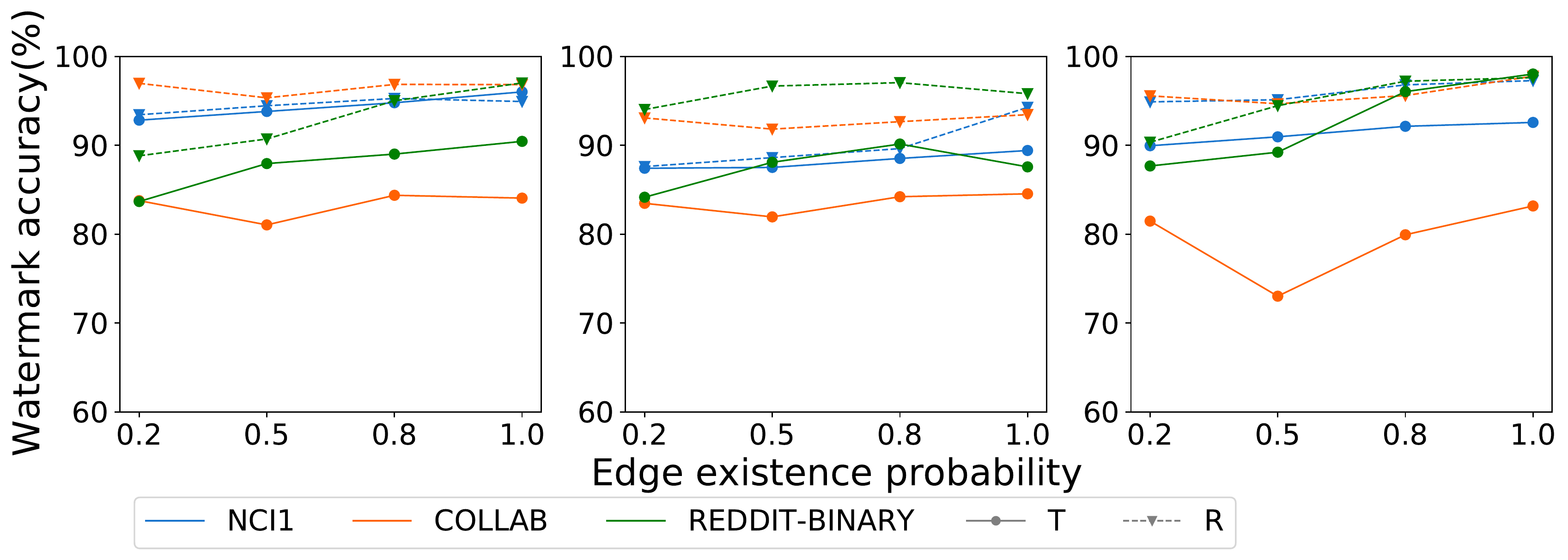}
         \caption{with different edge existence probability $p$ ($r=0.15, \gamma=0.2$)}
         \label{fig:asr_graph_classification-c}
     \end{subfigure}
        \caption{Watermark accuracy on graph classification task (DiffPool (left), GIN (center), GraphSAGE (right), T: first watermarked data generation strategy, R: second watermarked data generation strategy).}
        \label{fig:asr_graph_classification}
\end{figure}

\textbf{Graph classification.} As discussed in Section~\ref{watermark_generation}, there are two strategies for generating watermarked data $D_{wm}$. For each strategy, three parameters (watermarking rate $r$, watermark graph size $n$, and watermark graph density $p$) will affect the generated watermarked data and the final watermarking performance. The watermark accuracy of different datasets and models with different variants ($r, n, p$) is shown in Figure~\ref{fig:asr_graph_classification}. 
As we can see in Figure~\ref{fig:asr_graph_classification}, the watermark accuracy of the second watermarked data generation strategy is generally higher than the first strategy, which means the watermark pattern in the random graphs is more likely to be successfully learned by the model. 
In the first strategy, the original feature pattern in the graph may influence the learning of the watermark pattern, whereas, in the second strategy, embedding the watermark in random graphs can reduce this influence as it is the only important feature in the graph. 

From Figure~\ref{fig:asr_graph_classification-a}, we can see that for both datasets, with the increase of the watermarking rate, the watermark accuracy of all three models based on the first strategy is generally increasing, as well as for the second strategy.
Indeed, with a higher watermarking rate, more training data will be embedded with the watermark so that the host model can learn the watermark pattern better. However, we can also notice that even with the lowest watermarking rate, the watermark accuracy on all models and datasets is higher than the threshold in Table~\ref{watermark_acc_threshold_graph}, which indicates that the model owner can use a very small watermarking rate, e.g., $0.01$ to watermark his/her models. 
From Figure~\ref{fig:asr_graph_classification-b}, we can observe that in terms of watermark graph size from $\gamma=0.1$ to $\gamma=0.20$, the watermark accuracy of all three models and datasets gradually increases and then there is no significant increase (even slight decrease in some cases, e.g., REDDIT-BINARY) from $\gamma=0.20$ to $\gamma=0.25$. 
When the watermark graph gets larger, it is intuitive that the model can learn the watermark pattern easier and better. However, with continuous growth in the size of the watermark graph, there may not be enough model capacity to learn the watermark pattern.
In addition, from Figure~\ref{fig:asr_graph_classification-c}, for the NCI1 and REDDIT-BINARY datasets, the watermark accuracy grows slightly with the increase in the edge existence probability of the watermark graph. However, for the COLLAB dataset, the watermark accuracy first decreases for the range $p=0.2$ to $p=0.5$ and then increases.
The reason may be that when the watermark graph density is farther away from the graph density of the dataset, the trained model is more likely to recognize the watermark graph successfully. Therefore, the watermark accuracy is the lowest when $p=0.5$ for the COLLAB dataset, which has a density of $0.5089$. 
Furthermore, there is no apparent increase for the other two datasets, which have a density of $0.0889$ and $0.0218$, respectively. 


Based on the analysis of the results in Figure~\ref{fig:asr_graph_classification} and the later experimental results about the impact of watermarking GNNs on the original task, we set the parameters for the graph classification task as follows: $r=0.15, \gamma=0.2, p=1.0$. Specifically, Table~\ref{Table:asr_graph_classification} shows the watermark accuracy of model $m_w$ for the graph classification task with the selected parameters. 
For the binary-class datasets (i.e., NCI1 and REDDIT-BINARY), the accuracy on $D_{wm}^{t}$ is around $90\%$ while that of COLLAB is around $80\%$. This can be explained that COLLAB is a multi-class dataset, and it requires more model capacity to learn the features of each class so that the model has fewer redundant neurons to learn the watermark pattern compared to the other datasets. The watermark accuracy on $D_{wm}^{r}$ can mostly reach around $95\%$ for all datasets. 

\begin{table}
 \caption{Watermark accuracy for graph classification task ($r=0.15, \gamma=0.2, p=1.0$).}
\begin{center}
\begin{tabular}{cccc} 
 \hline
 \multirow{2}{*}{Dataset} & \multicolumn{3}{c}{Watermark Accuracy (\%) ($D_{wm}^{t}$ $|$ $D_{wm}^{r}$)}\\
 \cline{2-4}
 & {DiffPool} & {GIN} & {GraphSAGE} \\
 \hline
 \hline
 NCI1 & $96.01 | 94.92$ & $89.41 | 94.27$ & $92.57 | 97.28$\\
 \hline
 COLLAB & $84.05 | 96.83$ & $84.54 | 93.45$ & $83.17 | 97.77$\\
 \hline 
 REDDIT-BINARY & $90.43 | 97.01$ & $87.57 | 95.81$ & $98.02 | 97.61$ \\
 \hline
 \end{tabular}
\label{Table:asr_graph_classification}
\end{center}
\end{table}

\begin{figure}[!ht]
     \centering
     \begin{subfigure}{0.48\textwidth}
         \centering
         \includegraphics[width=1\textwidth]{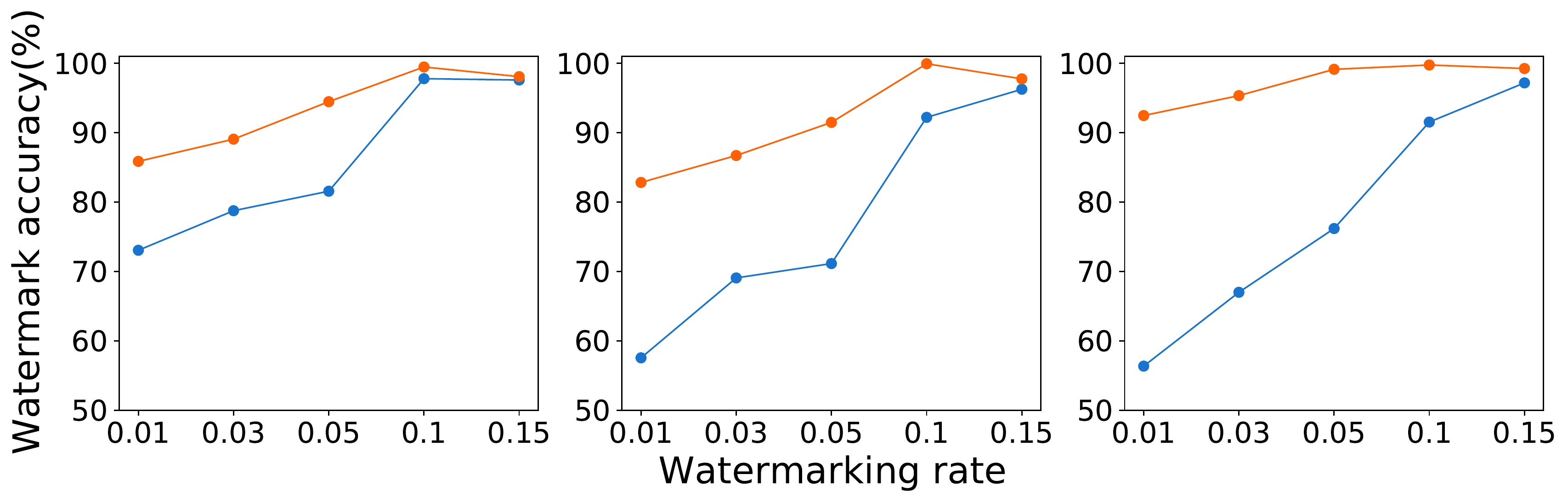}
         \caption{with the different watermarking rate $r$}
         \label{fig:asr_node_classification-a}
     \end{subfigure}
     \newline
     \begin{subfigure}{0.48\textwidth}
         \centering
         \includegraphics[width=1\textwidth]{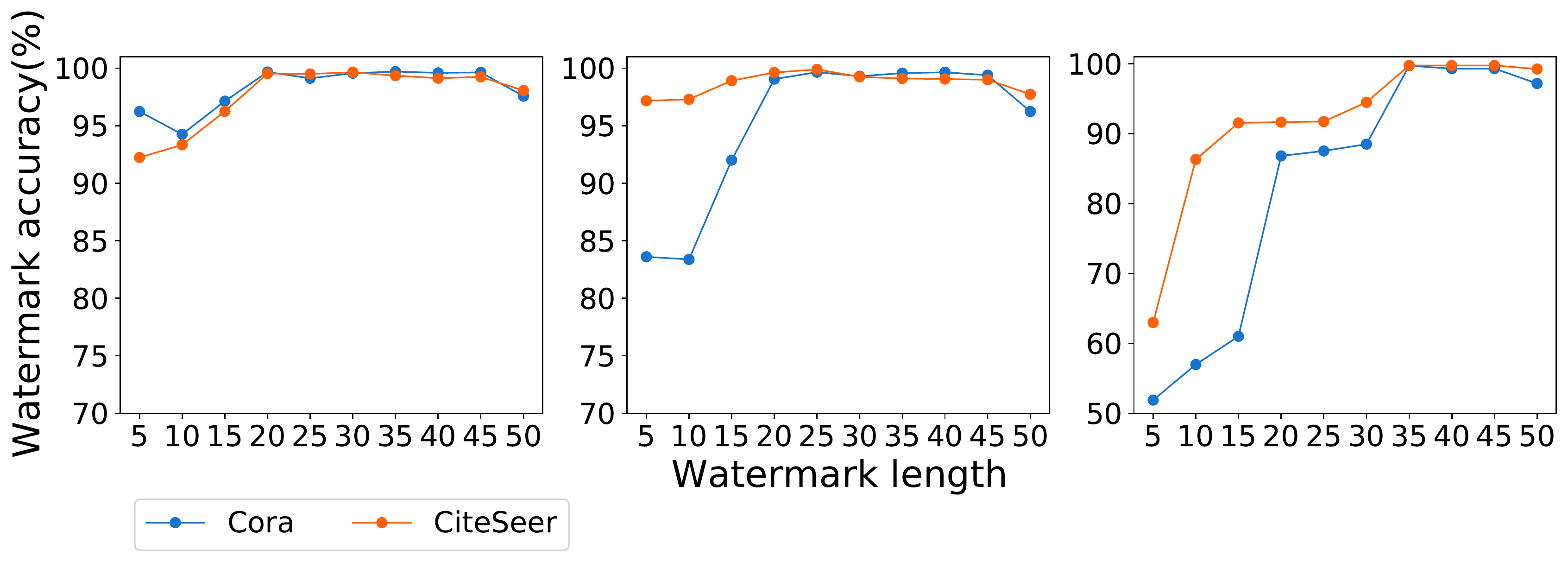}
         \caption{with different watermark length $l$}
         \label{fig:asr_node_classification-b}
     \end{subfigure}
        \caption{Watermark accuracy on node classification task (GCN (left), GAT (center), GraphSAGE (right)).}
        \label{fig:asr_node_classification}
\end{figure}

\begin{table}[!ht]
 \caption{Watermark accuracy for the node classification task.}
\begin{center}
\begin{tabular}{cccc}
 \hline
 \multirow{2}{*}{Dataset} & \multicolumn{3}{c}{Watermark Accuracy (\%)} \\
 \cline{2-4}
 & {GCN} & {GAT} & {GraphSAGE}\\
 \hline
 \hline
 Cora & $97.56$ & $96.23$ & $97.15$ \\
 \hline
 CiteSeer & $98.05$ & $97.73$ & $99.22$\\
 \hline
\end{tabular}
\label{Table:asr_node_classification}
\end{center}
\end{table}

\begin{figure}[!ht]
     \centering
     \begin{subfigure}{0.48\textwidth}
         \centering
         \includegraphics[width=1\textwidth]{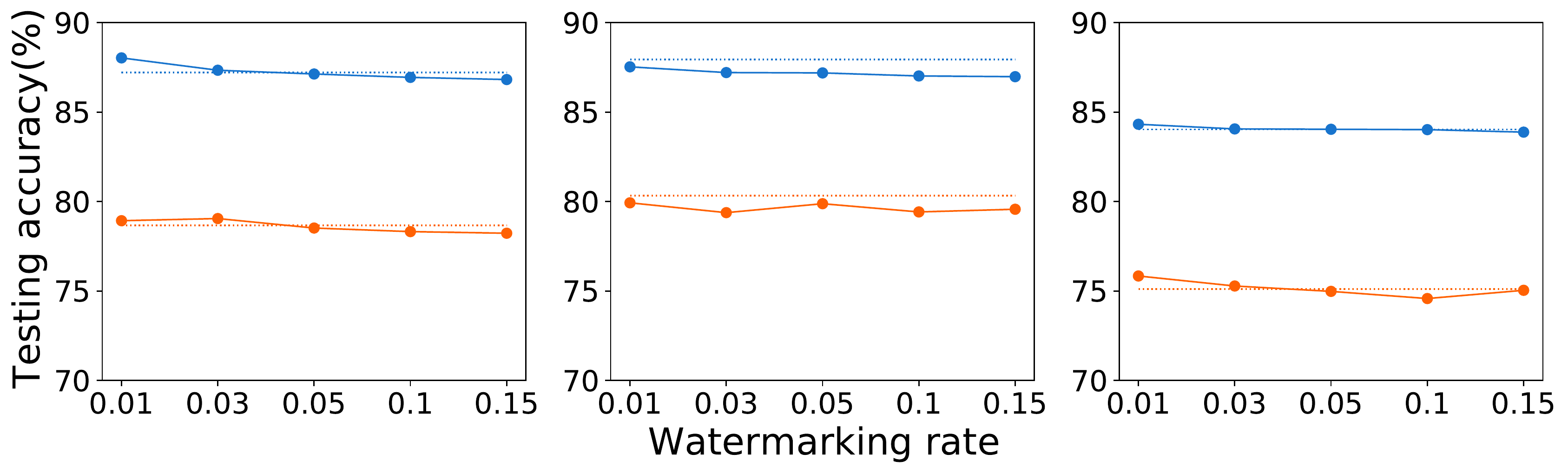}
         \caption{with the different watermarking rate $r$}
         \label{fig:test_acc_node_classification-a}
     \end{subfigure}
     \newline
     \begin{subfigure}{0.48\textwidth}
         \centering
         \includegraphics[width=1\textwidth]{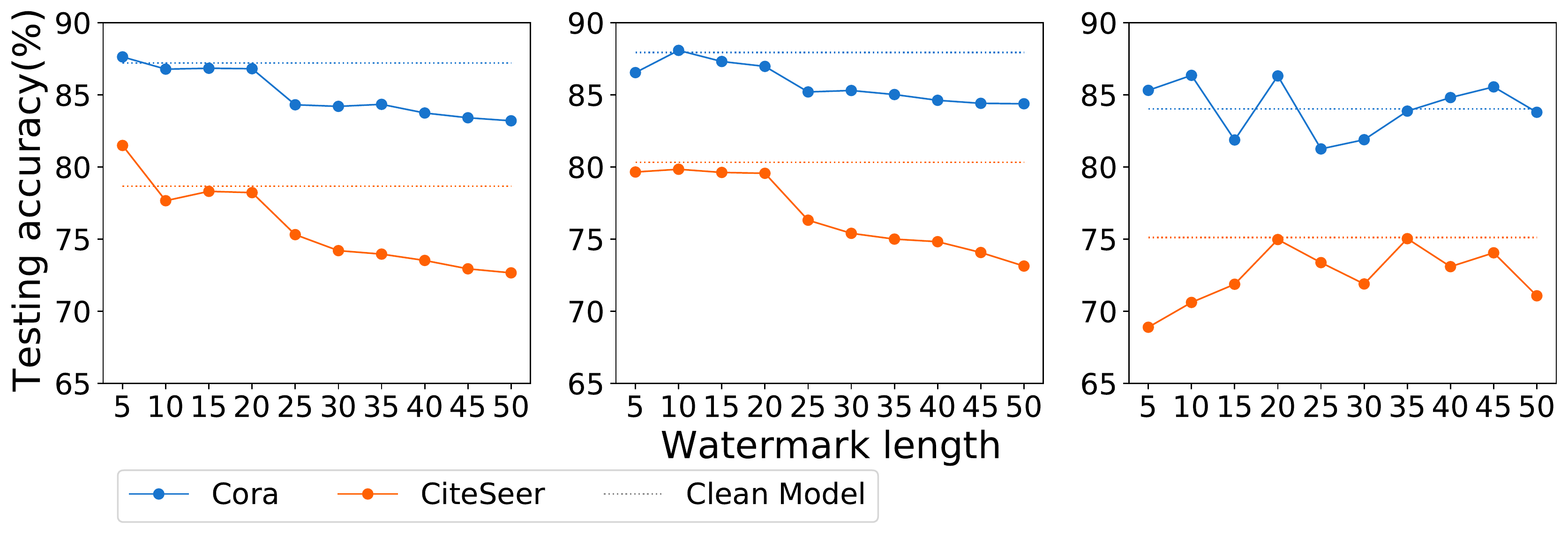}
         \caption{with different watermark length $l$}
         \label{fig:test_acc_node_classification-b}
     \end{subfigure}
        \caption{Testing accuracy on node classification task (GCN (left), GAT (center), GraphSAGE (right)).}
        \label{fig:test_acc_node_classification}
\end{figure}

\textbf{Node classification.} For the node classification task, the generated watermarked data is decided by two parameters (watermarking rate $r$ and feature watermark length $l$). The watermark accuracy of three GNN models, i.e., GCN, GAT, and GraphSAGE, for the node classification task is shown in Figure~\ref{fig:asr_node_classification}. As we can see from Figure~\ref{fig:asr_node_classification-a}, for all datasets and models, the watermark accuracy has a dramatic rise with $r$ ranging from $0.01$ to $0.1$, and slight increase when $r$ is in the range of $\left [0.1, 0.15\right ]$.
From Figure~\ref{fig:asr_node_classification-b}, it can be observed that for the GCN and GAT models, there is a significant increase between $l=5$ and $l=20$ and when $l$ continues rising to $l=50$, there is no obvious effect for both datasets. For the GraphSAGE model, the watermark accuracy gradually increases from $l=5$ to $l=35$ and stays steady. 
This is expected since, with more nodes embedded in the feature watermark, the GNN model can learn the watermark pattern better. Also, the GNN model can better memorize the watermark pattern with a larger watermark. 
Moreover, for watermarks with a watermark length of less than $35$, the watermark accuracy for GCN and GAT models is higher than the GraphSAGE model indicating that GCN and GAT models learn small watermarks better than the GraphSAGE model. 
Since transductive learning has the advantage of being able to directly use training patterns while deciding on a test pattern~\cite{bianchini2016comparative}, it is easier for the GCN and GAT models (under transductive learning setting) to learn the watermark pattern of specific size than for the GraphSAGE model (under inductive learning setting).
Considering the results in Figures~\ref{fig:asr_node_classification} and~\ref{fig:test_acc_node_classification} (which will be further analyzed later), we set the parameters for the node classification task as follows: $r=0.15, l=20$ for the GCN and GAT models, and $r=0.15, l=35$ for the GraphSAGE model. Table~\ref{Table:asr_node_classification} shows the watermark accuracy of three models for the node classification task with the selected parameters. 

\begin{figure}[!ht]
    \centering
    \includegraphics[width=0.33\textwidth]{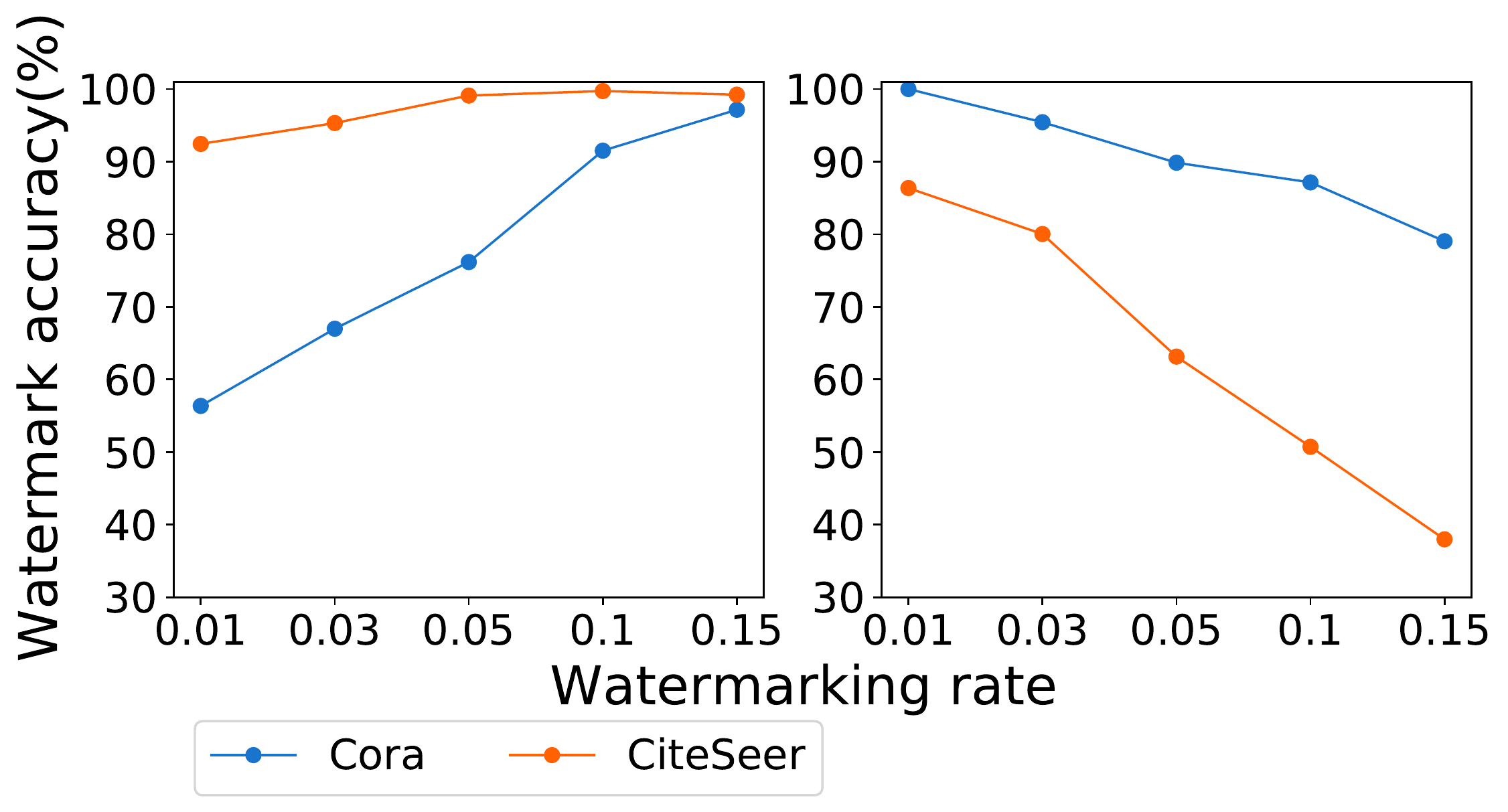}
    \caption{Comparison between our method (left) and~\cite{zhao2021watermarking} (right).}
    \label{fig:node_comparison}
\end{figure}

We also compare our watermarking mechanism on the node classification task with the state-of-the-art~\cite{zhao2021watermarking}, as shown in Figure~\ref{fig:node_comparison}. 
As we can see, with the increasing watermarking rate, the watermark accuracy in that work declines to $79\%$ and $38\%$ for Cora and CiteSeer, respectively. Contrarily, in our watermarking method, the watermark accuracy keeps increasing to $100\%$ for both datasets. The decline of watermark accuracy is consistent and explained in~\cite{zhao2021watermarking}. 
It can be seen that our watermarking method can achieve similar or higher watermark accuracy than the state-of-the-art.

\begin{figure}[!ht]
     \centering
     \begin{subfigure}{0.48\textwidth}
         \centering
         \includegraphics[width=\textwidth]{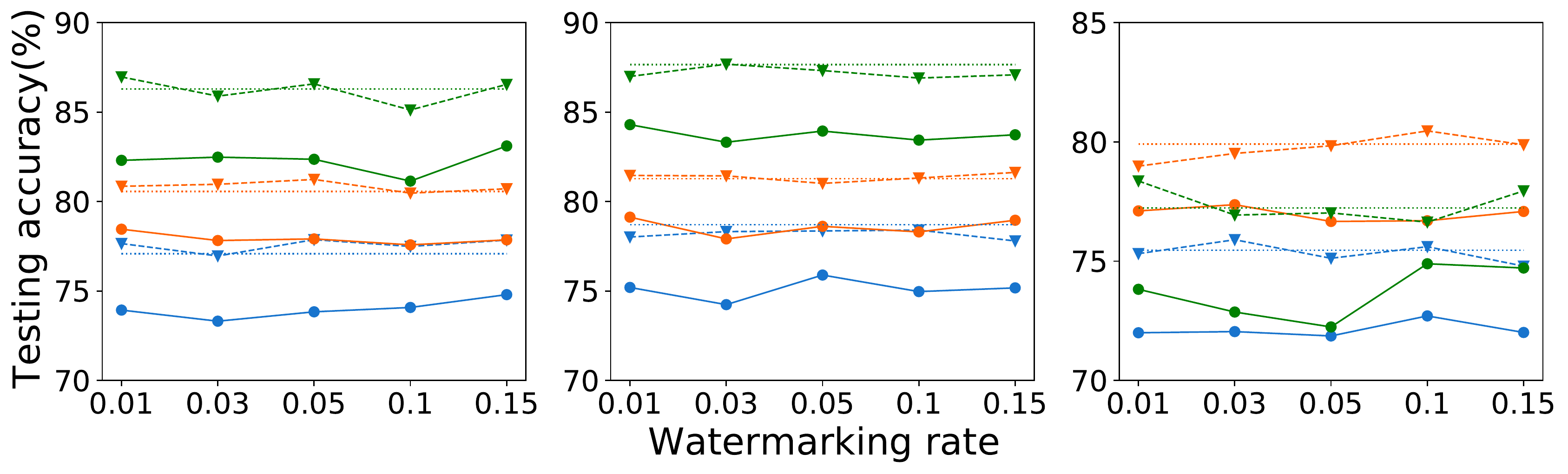}
         \caption{with different watermarking rate $r$ ($\gamma=0.2, p=1.0$)}
         \label{fig:test_acc_graph_classification-a}
     \end{subfigure}
     \newline
     \begin{subfigure}{0.48\textwidth}
         \centering
         \includegraphics[width=\textwidth]{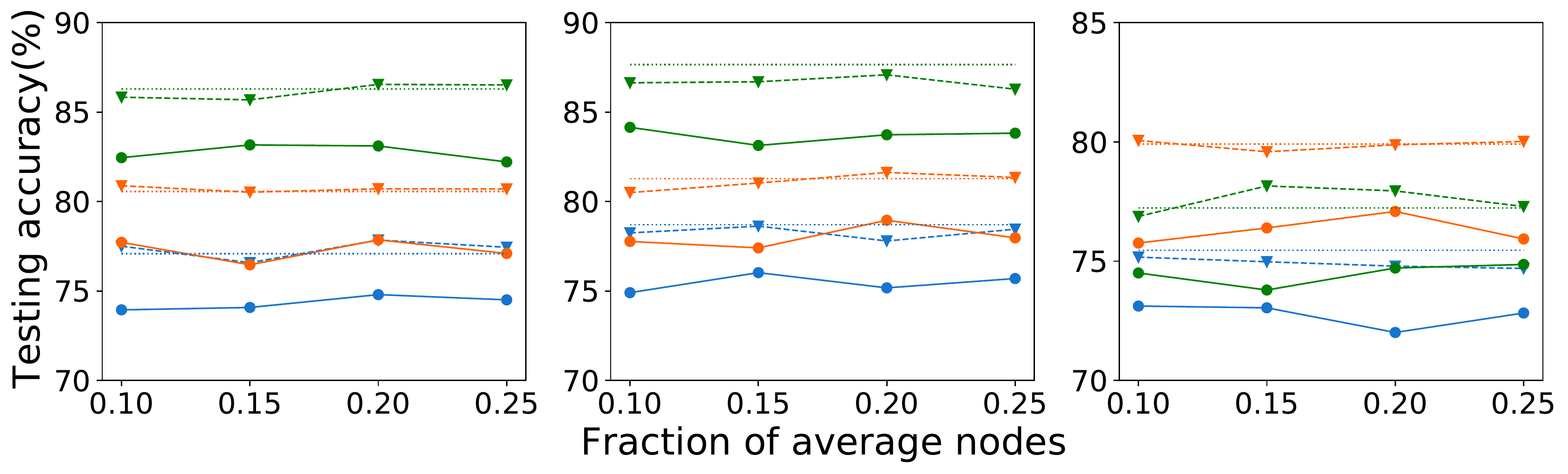}
         \caption{with different fraction of average nodes $\gamma$ ($r=0.15, p=1.0$)}
         \label{fig:test_acc_graph_classification-b}
     \end{subfigure}
     \newline
     \begin{subfigure}{0.48\textwidth}
         \centering
         \includegraphics[width=\textwidth]{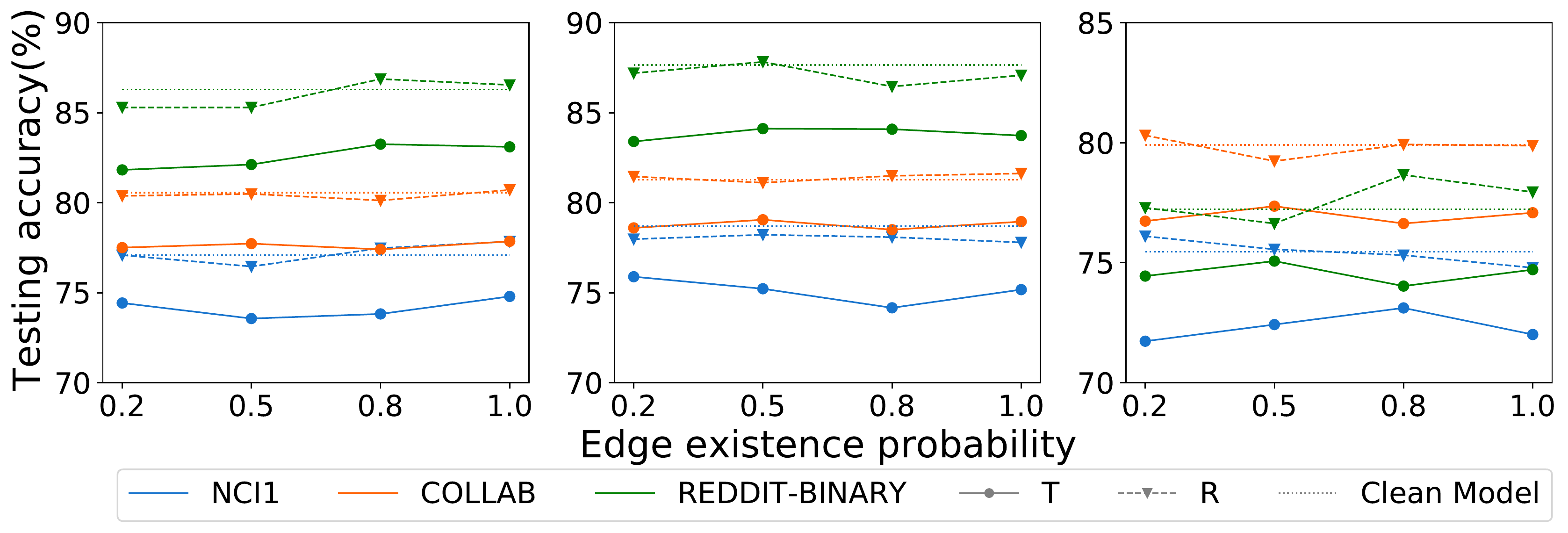}
         \caption{with different edge existence probability $p$ ($r=0.15, \gamma=0.2$)}
         \label{fig:test_acc_graph_classification-c}
     \end{subfigure}
        \caption{Testing accuracy on graph classification task (DiffPool (left), GIN (center), GraphSAGE (right), T: first watermarked data generation strategy, R: second watermarked data generation strategy).}
        \label{fig:test_acc_graph_classification}
\end{figure}

\textbf{Impact on the original task.} To measure the impact of our watermarking mechanism on the watermarked model's original task, we measure the accuracy of models on the normal test data. The test data is not used to train the host model. Figures~\ref{fig:test_acc_graph_classification} and~\ref{fig:test_acc_node_classification} illustrate the testing accuracy of models with and without embedding watermarks under different variants for graph and node classification tasks, respectively. 
We use the testing accuracy of the clean model as the baseline, i.e., if the testing accuracy of the watermarked model is close to the baseline, we can confirm that our watermarking mechanism will not affect the watermarked model's original task. 
We can see from Figure~\ref{fig:test_acc_graph_classification} that, for all three models and datasets, the testing accuracy of the second watermarked data generation strategy is always much closer to the baseline than the first strategy, which means the second strategy has less impact on the model's original task.
In the first strategy, the watermarked data is generated by embedding a watermark into sampled training data, so it is possible to embed a watermark into a graph structure, which has a critical effect on the final prediction. As a result, the watermarking process of the first strategy will probably affect the parameters in the networks that are used for the original task. However, in the second strategy, random graphs are used as the watermark carrier data, so the model will try to explore redundancy in the network capacity to learn the watermark pattern while not affecting the original task.
 
For the node classification task (Figure~\ref{fig:test_acc_node_classification}), the watermarking rate has a negligible impact on the testing accuracy, while when the watermark length is larger than $20$, there is a significant reduction in the testing accuracy for GCN and GAT models. 
For the GraphSAGE model, the testing accuracy fluctuates with the increase in watermark length. 
Referring to the watermark accuracy in Figure~\ref{fig:asr_node_classification-b}, one possible reason is that if the watermark size continuously increases, the model's redundant capacity will be fully occupied, and the model will use some neurons originally for the main task to keep high watermark accuracy. Thus, the testing accuracy for GCN and GAT decreases when the watermark length is larger than $20$. For GraphSAGE, the testing accuracy does not reduce significantly because the GraphSAGE model has more redundant neurons than the other two models. 
Tables~\ref{Table:test_acc_graph_classification} and~\ref{Table:test_acc_node_classification} show the testing accuracy with the selected parameters for graph and node classification tasks, respectively. As we can see for the graph classification task, the testing accuracy of the first strategy is about $3\%$ less than that of the second strategy. There is less than $1\%$ clean accuracy drop for the node classification task.

\begin{table}[!ht]
 \caption{Testing accuracy for graph classification task ($r=0.15, \gamma=0.2, p=1.0$).}
\begin{center}
\begin{subtable}[t]{0.48\textwidth}
\centering
\caption{\footnotesize NCI1}
\begin{tabular}{cccc}
 \hline
{Testing Accuracy (\%)} & {DiffPool} & {GIN} & {GraphSAGE} \\
 \hline
 \hline
 {Clean Model} & $77.09$ & $78.71$ & $75.46$\\
 \hline
 $D_{wm}^{t}$ & $74.80$ & $75.18$ & $72.01$\\
 \hline
 $D_{wm}^{r}$ & $77.85$ & $77.80$ & $74.79$\\
 \hline
\end{tabular}
\label{Table:test_acc_graph_classification_a}
\end{subtable}

\begin{subtable}[t]{0.48\textwidth}
\centering
\caption{\footnotesize COLLAB}
\begin{tabular}{cccc}
 \hline
{Testing Accuracy (\%)} & {DiffPool} & {GIN} & {GraphSAGE} \\
 \hline
 \hline
 {Clean Model} & $80.56$ & $81.27$ & $79.91$\\
 \hline
 $D_{wm}^{t}$ & $77.86$ & $78.96$ & $77.09$\\
 \cline{1-4}
 $D_{wm}^{r}$ & $80.71$ & $81.63$ & $79.88$\\
 \hline
\end{tabular}
\label{Table:test_acc_graph_classification_b}
\end{subtable}

\begin{subtable}[t]{0.48\textwidth}
\centering
\caption{\footnotesize REDDIT-BINARY}
\begin{tabular}{cccc}
 \hline
{Testing Accuracy (\%)} & {DiffPool} & {GIN} & {GraphSAGE} \\
 \hline
 \hline
 {Clean Model} & $86.30$ & $87.65$ & $77.23$\\
 \hline
 $D_{wm}^{t}$ & $83.11$ & $83.73$ & $74.71$\\
 \cline{1-4}
 $D_{wm}^{r}$ & $86.55$ & $87.08$ & $77.95$\\
 \hline
\end{tabular}
\label{Table:test_acc_graph_classification_c}
\end{subtable}
\label{Table:test_acc_graph_classification}
\end{center}
\end{table}


\begin{table}
 \caption{Testing accuracy for the node classification task.}
\begin{center}
\begin{tabular}{cccc} 
 \hline
 \multirow{2}{*}{Dataset} & \multicolumn{3}{c}{Testing Accuracy (\%) ($Clean Model | D_{wm}^{t}$)} \\
 \cline{2-4}
  & {GCN} & {GAT} & {GraphSAGE} \\
 \hline
 \hline
 Cora & $87.21 | 86.82$ & $87.94 | 86.98$ & $84.04 | 83.88$\\
 \hline
 CiteSeer & $78.67 | 78.23$ & $80.33 | 79.57$ & $75.12 | 75.04$\\
 \hline 
 \end{tabular}
\label{Table:test_acc_node_classification}
\end{center}
\end{table}

\subsection{On the Watermarking Requirements}
\label{sec:watermark_requirements}

In this section, we explain the well-known watermarking requirements for neural networks given in~\cite{DBLP:journals/ijon/LiWB21,DBLP:journals/fdata/Boenisch21} and how our watermarking framework achieves them.

\textbf{Robustness.}
It refers to the resistance of the watermarking against the modifications that can be caused by malicious perturbations\footnote{In~\cite{DBLP:journals/ijon/LiWB21}, this is separately defined as the \textit{security} requirement.} or benign processing.
A robust GNN watermarking should be recoverable even after the model has been modified by fine-tuning and network pruning.
As we discussed in Section~\ref{sec:robustness}, our watermarking model achieves robustness against the modifications in the model.

\textbf{Fidelity.}
It is about maintaining the \textit{quality} of the watermarked object while watermarking.
A GNN watermarking model satisfies the fidelity requirement if it does not significantly degrade the performance of the GNN model.
Our experimental results presented in Tables~\ref{Table:test_acc_graph_classification} and~\ref{Table:test_acc_node_classification} show that our watermarking (with random graphs) on graph classification and on node classification tasks leads to less than 1\% decrease on the accuracy of the original task. 
Detailed discussion is given in Section~\ref{results}.

\textbf{Capacity.}
It is defined as the watermark's capability to carry information, and two watermarking schemes, zero-bit, and multi-bit, are distinguished by it~\cite{boenisch2021systematic}. Our work applies zero-bit watermarks because they do not carry additional information, such that they solely serve to indicate the presence or the absence of the watermark in a model. 

\textbf{Integrity.}
It refers to the correct classification of the watermarking.
A GNN watermarking mechanism satisfying integrity should have low FPR and FNR.\footnote{In~\cite{DBLP:journals/ijon/LiWB21}, the false negative case is defined as the \textit{reliability} requirement.}
Here, the false positive requirement states that a benign model not copied from the watermarked GNN model should not be seen as a malicious copy.
Whereas the false negative implies that a malicious copy of the watermarked GNN model should be classified as a copy. 
As mentioned in Sections~\ref{subsec:ownership_verification} and~\ref{sec:experiments}, we select a watermark accuracy threshold for each dataset and model to make sure the FPR and FNR are less than $0.0001$. 


\textbf{Generality.}
It is about the generalization of the watermarking method.
A GNN watermarking would have generality if it is not tailored to a specific model but can be applied to other architectures or models. Our watermarking method has been evaluated in several state-of-the-art GNN models. In addition, our watermarking method does not depend on the architectures of the GNN models and can be applied to any other GNN models and graph datasets.

\textbf{Efficiency.}
It concerns the overhead required for embedding and verifying a watermark into an object.
An efficient GNN watermarking model should not add too much computational cost by adding the watermark. In our watermarking method, according to the watermark embedding algorithm~\ref{alg:watermark_embedding} and the watermarking rate we set, the overhead in embedding watermark into a GNN model is retraining a clean GNN model with $30\%$ or $15\%$ of the original training data. 
The overhead in verifying the watermark is the number of watermarked data which is related to the watermarking rate, i.e., between $0.01$ and $0.15$. 
As we explained in Section~\ref{results}, in our watermarking method, the model owners can use small watermarking rates, e.g., $0.01$, to watermark their models. 

Moreover, in~\cite{DBLP:journals/fdata/Boenisch21}, the authors also defined \textit{secrecy} property that is about the detectability of the presence of the watermark.
However, this property is out of scope for our watermarking framework because we expect our watermarking mechanism still works even if the adversaries know the existence of the watermark.\footnote{But the watermark itself should be secret to the adversaries.} 
\section{Robustness Against Backdoor Defenses}
\label{sec:robustness}

Our watermarking method is based on backdoor attacks, so it is intuitive to explore whether it is resistant to state-of-the-art defenses against backdoor attacks. The state-of-the-art defenses against backdoor attacks can be summarized in four categories: input reformation, input filtering, model sanitization, and model inspection~\cite{pang2020trojanzoo}. NeuralCleanse (NC)~\cite{wang2019neural} is the most representative defense in the model inspection defense category. However, it is not feasible to be applied in this work because (1) NC requires a large number of input samples to achieve good performance~\cite{liu2019abs} while in our work, the plagiarizer has no access to the training data, and (2) NC cannot reverse engineer large triggers~\cite{februus,liu2019abs,meta-neural-analysis}, while in our work, ideally, there is no restriction on the watermark size.
Similarly, as one of the most representative defenses in the input filtering defense category, Activation-Clustering (AC)~\cite{chen2018detecting} is not applicable because it requires access to poisoned training data to remove the backdoor, but in our setting, the plagiarizer has no knowledge of the watermarked data.
On the other hand, Randomized-Smoothing (RS)~\cite{cohen2019certified} (input reformation defense) can be applied because it can only reform the input samples without the requirement of knowledge of watermarked data. Three model modifications (i.e., fine-tuning, model pruning, and fine-pruning), which can be categorized into the model sanitization defenses, are also applied in our work to explore the robustness of our watermarked model.

Next, we investigate the robustness of our watermarked model against a state-of-the-art model extraction technique: knowledge distillation, and state-of-the-art defenses against backdoor attacks: randomized subsampling, fine-tuning, pruning, and fine-pruning.

\textbf{Robustness against knowledge distillation.}
Knowledge distillation aims at transferring knowledge from a teacher model to a student~\cite{deng2021graph}. It has been used in the model extraction attacks where the teacher model is the victim model, and the student model is the stolen extracted model~\cite{truong2021data}. Here, we suppose the plagiarizer applies knowledge distillation to extract the knowledge from the host model to train the plagiarized model, and we explore the robustness of our watermarking method against this attack. Specifically, we follow the offline distillation strategy in~\cite{gou2021knowledge} since the host model is pre-trained. We assume the student model has the same model structure as the teacher model, and we use half of the test data to be the training data for the knowledge distillation.\footnote{We assume that the plagiarizer has access to the testing data aiming to explore the robustness of our method under a strong adversary.} We evaluate the distilled model, i.e., the student model, with the second half of the test data. 
Tables~\ref{Table:asr_kd_graph_classification} and~\ref{Table:test_acc_kd_graph_classification} show the watermark accuracy and testing accuracy of the model after knowledge distillation on the graph classification task. We can observe that the watermark accuracy decrease is less than $3.25\%$ for two watermarked data generation strategies, for all datasets and models, except $4.99\%$ decline for the NCI1 dataset with the GIN model. As for the testing accuracy, we see that knowledge distillation has little impact on the model's original task. For the node classification task, the watermarking performance after the knowledge distillation is shown in Table~\ref{Table:kd_node_classification}. As we can see, the watermark accuracy after the knowledge distillation even increases a little on the node classification task. 
Since the knowledge distillation can improve the generalization of the student model~\cite{stanton2021does}, it may reduce the overfitting in the original model.
Moreover, the testing accuracy decreases negligibly except for CiteSeer with the GAT and GraphSAGE models. One possible explanation is that in our work, the GAT and GraphSAGE models are more complex than GCN, so it is more difficult for these two models to transfer the knowledge completely from the teacher model to the student model. 
Based on the observations above, we can claim that knowledge distillation can transfer the knowledge of the host model on the original task to the student model successfully. Still, it can also transfer the watermarking function, which means our watermarking mechanism is robust against knowledge distillation.

\begin{table}[!htpb]
\begin{center}
 \caption{Accuracy on watermarked data after knowledge distillation for graph classification task ($r=0.15, \gamma=0.2, p=1.0$).}
\begin{tabular}{cccc} 
 \hline
 \multirow{2}{*}{Dataset} & \multicolumn{3}{c}{Watermark Accuracy (\%) ($D_{wm}^{t} | D_{wm}^{r}$)} \\
 \cline{2-4}
 & {DiffPool} & {GIN} & {GraphSAGE} \\
 \hline
 \hline
NCI1 & $92.99 | 93.88$ & $87.90 | 89.28$ & $90.36 | 96.27$ \\
\hline
COLLAB & $83.11 | 95.87$ & $82.66 | 92.75$ & $80.77 | 94.54$ \\
\hline
REDDIT-BINARY & $87.74 | 96.92$ & $89.71 | 97.45$ & $97.53 | 97.67$ \\
\hline
 \end{tabular}
\label{Table:asr_kd_graph_classification}
\end{center}
\end{table}

\begin{table}[htpb]
 \caption{Testing accuracy after knowledge distillation for graph classification task ($r=0.15, \gamma=0.2, p=1.0$).}
\begin{center}
\begin{tabular}{cccc} 
 \hline
 \multirow{2}{*}{Dataset} & \multicolumn{3}{c}{Testing Accuracy (\%) ($D_{wm}^{t} | D_{wm}^{r}$)} \\
 \cline{2-4}
  & {DiffPool} & {GIN} & {GraphSAGE} \\
 \hline
 \hline
 NCI1 & $75.92 | 77.30$ & $78.50 | 78.88$ & $76.83 | 76.46$ \\
 \hline
 COLLAB & $76.18 | 80.79$ & $77.67 | 81.09$ & $75.08 | 79.60$\\
 \hline 
 REDDIT-BINARY & $83.63 | 86.35$ & $83.82 | 87.54$ & $77.65 | 78.25$\\
 \hline
 \end{tabular}
\label{Table:test_acc_kd_graph_classification}
\end{center}
\end{table}

\begin{table}[!htpb]
 \caption{Watermarking performance after knowledge distillation on the node classification task.}
\begin{center}
\begin{tabular}{cccc} 
 \hline
 \multirow{2}{*}{Dataset} & \multicolumn{3}{c}{Watermark Accuracy (\%) $|$ Testing Accuracy (\%)} \\
 \cline{2-4}
  & {GCN} & {GAT} & {GraphSAGE} \\
 \hline
 \hline
Cora & $99.64 | 88.22$ & $98.98 | 86.28$ & $99.23 | 82.19$ \\
\hline
CiteSeer & $99.51 | 78.26$ & $99.59 | 73.65$ & $99.30 | 73.54$ \\
\hline
 \end{tabular}
\label{Table:kd_node_classification}
\end{center}
\end{table}

\textbf{Robustness against fine-tuning.} As discussed before, training a well-performed GNN model from scratch requires a large amount of training data and powerful computing resources. Fine-tuning is a practical attack on GNNs since it can be used to apply existing state-of-the-art models to other but similar tasks with less effort than training a network from scratch when sufficient training data is not available~\cite{yosinski2014transferable}. 
Therefore, fine-tuning is likely to be used by a suspect to train a new model on top of the stolen model using only a small amount of training data.

In this experiment, for each dataset, we use half of the test data to fine-tune the previously trained watermarked GNN model, and the second half is used to evaluate the new model. Then, we use the watermark accuracy to determine whether the watermark embedded in the previously trained GNN model stayed effective in the new model. Additionally, the testing accuracy is used to evaluate the performance of the newly trained model on its original task. Tables~\ref{Table:asr_fine_tuning_graph_classification} and~\ref{Table:test_acc_fine_tuning_graph_classification} show the watermark accuracy and testing accuracy of the model after fine-tuning for graph classification task. Comparing the results from Tables~\ref{Table:asr_graph_classification} and~\ref{Table:asr_fine_tuning_graph_classification}, we can see that fine-tuning does not  significantly reduce  (less than $4.65\%$) the watermark accuracy for all datasets and models. 
Triggers rarely appear in the fine-tuning dataset; consequently, the backdoor functionality will not be eliminated~\cite{zhang2021red}.
As for the testing accuracy, from Tables~\ref{Table:test_acc_graph_classification} and~\ref{Table:test_acc_fine_tuning_graph_classification}, we can observe that after fine-tuning, the embedded watermark in the model still has only a slight effect on the model's original task.
For the node classification task, fine-tuning is not feasible for transductive learning-based GNN models because once the training data change, the model should be retrained from scratch. Therefore, we show here the results of the GraphSAGE model (inductive learning) for the node classification task, as shown in Table~\ref{Table:fine_tunning_node_classification}, and the observations are similar to that for the graph classification task.

\begin{table}[!htpb]
\begin{center}
 \caption{Watermark accuracy after fine-tuning for graph classification task ($r=0.15, \gamma=0.2, p=1.0$).}
\begin{tabular}{cccc} 
 \hline
 \multirow{2}{*}{Dataset} & \multicolumn{3}{c}{Watermark Accuracy (\%) ($D_{wm}^{t} | D_{wm}^{r}$)} \\
 \cline{2-4}
 & {DiffPool} & {GIN} & {GraphSAGE} \\
 \hline
 \hline
NCI1 & $94.00 | 93.69$ & $88.71 | 89.64$ & $91.92 | 97.17$ \\
\hline
COLLAB & $83.71 | 96.50$ & $82.96 | 91.83$ & $80.48 | 94.11$ \\
\hline
REDDIT-BINARY & $87.61 | 96.78$ & $90.11 | 97.87$ & $98.68 | 97.90$ \\
\hline
 \end{tabular}
\label{Table:asr_fine_tuning_graph_classification}
\end{center}
\end{table}

\begin{table}[!htpb]
 \caption{Testing accuracy after fine-tuning for graph classification task ($r=0.15, \gamma=0.2, p=1.0$).}
\begin{center}
\begin{tabular}{cccc} 
 \hline
 \multirow{2}{*}{Dataset} & \multicolumn{3}{c}{Testing Accuracy (\%) ($D_{wm}^{t} | D_{wm}^{r}$)} \\
 \cline{2-4}
  & {DiffPool} & {GIN} & {GraphSAGE} \\
 \hline
 \hline
 NCI1 & $73.61 | 77.15$ & $75.05 | 77.84$ & $71.92 | 74.89$ \\
 \hline
 COLLAB & $76.10 | 80.67$ & $77.63 | 80.92$ & $75.93 | 78.31$\\
 \hline 
 REDDIT-BINARY & $82.10 | 86.20$ & $80.63 | 86.89$ & $73.82 | 77.85$\\
 \hline
 \end{tabular}
\label{Table:test_acc_fine_tuning_graph_classification}
\end{center}
\end{table}

\begin{table} [!ht]
 \centering
 \caption{Watermarking performance of GraphSAGE model after fine-tuning on the node classification task.}
\begin{tabular}{ccc}
 \hline
 {Dataset} & {Watermark Accuracy (\%)} & {Testing accuracy (\%)} \\
 \hline
 \hline
 Cora & $98.51$ & $83.56$  \\
 \hline
 CiteSeer & $99.45$ & $75.10$ \\
 \hline
\end{tabular}
\label{Table:fine_tunning_node_classification}
\end{table}

\textbf{Robustness against model pruning.}
\label{model_pruning}
Model pruning is a technique to develop a neural network model that is smaller and more efficient by setting some parameters to zero while maintaining the performance on the primary task~\cite{zhang2018protecting}. We apply the pruning algorithm used in~\cite{uchida2017embedding}, which prunes the parameters whose absolute values are very small. Specifically, for all the watermarked models, we remove a certain number of parameters with the smallest absolute values by setting them to zero. The ratio between the number of pruned parameters and the total number of parameters is the pruning rate (here from $10\%$ to $90\%$). Then, we measure the watermark and the testing accuracy of the pruned watermarked model.

Tables~\ref{Table:pruning_NCI1} and~\ref{Table:pruning_Cora} present the watermarking performance after model pruning on the graph and node classification tasks, respectively. We take the results for NCI1 and Cora as examples (the results of other datasets are shown in Appendix~\ref{appendix:model_pruning}). For the NCI1 dataset, even when $40\%$ of the parameters are pruned, our watermarked model still has a high watermark accuracy, i.e., less than $1\%$ drop in all models. Especially for the GraphSAGE model, there is only a $0.68\%$ drop even though $70\%$ of the parameters are pruned. We can also observe that when $90\%$ of the parameters are pruned, the watermark accuracy drops dramatically for all models, e.g., it drops to less than $10\%$ for the DiffPool model. However, we also notice that in this case, there is a significant testing accuracy drop for the model (more than $20\%$), which means the plagiarizer has to take the expense of dramatically degrading the model's performance on the original task to eliminate our watermark. 
As for the Cora dataset, the watermark accuracy decreases gradually as well as the testing accuracy for GCN and GAT models. For the GraphSAGE model, model pruning also leads to a more obvious watermark accuracy drop, i.e., more than $60\%$, as well as an apparent testing accuracy drop (more than $10\%$). 
Thus, our watermarking mechanism is generally robust to model pruning. The plagiarizer can only eliminate our watermarks with the cost of a considerable accuracy drop in the main task.

\begin{table*}[!htpb]
\centering
 \caption{Watermarking performance on graph classification task after model pruning (NCI1).}
\begin{center}
\begin{tabular}{ccccccc}
 \hline
\multirow{2}{*}{Pruning} & \multicolumn{2}{c}{DiffPool} & \multicolumn{2}{c}{GIN} & \multicolumn{2}{c}{GraphSAGE} \\
\cline{2-7}
  & {Test Acc. } & {Watermark Acc.} & {Test Acc.} & {Watermark Acc.} & {Test Acc.} & {Watermark Acc.}\\
 {rate} & ($D_{wm}^{t} | D_{wm}^{r}$) & ($D_{wm}^{t} | D_{wm}^{r}$) & ($D_{wm}^{t} | D_{wm}^{r}$) & ($D_{wm}^{t} | D_{wm}^{r}$) & ($D_{wm}^{t} | D_{wm}^{r}$) & ($D_{wm}^{t} | D_{wm}^{r}$) \\
\hline
{$10\%$ } & {$77.77\%|77.86\%$} & {$95.52\%|95.40\%$} & {$78.21\%|77.81\%$} & {$89.96\%|90.38\%$} & {$74.66\%|74.83\%$} & {$93.06\%|97.65\%$} \\
{$20\%$ } & {$77.72\%|77.84\%$} & {$95.47\%|95.26\%$} & {$77.79\%|77.62\%$} & {$89.82\%|90.51\%$} & {$74.77\%|74.46\%$} & {$92.83\%|97.68\%$} \\
{$30\%$ } & {$77.61\%|77.69\%$} & {$94.70\%|95.48\%$} & {$77.24\%|76.95\%$} & {$89.41\%|90.63\%$} & {$73.56\%|73.93\%$} & {$92.78\%|97.55\%$} \\
{$40\%$ } & {$76.79\%|77.25\%$} & {$94.68\%|95.36\%$} & {$72.93\%|71.76\%$} & {$89.33\%|90.50\%$} & {$73.32\%|72.36\%$} & {$92.71\%|97.47\%$} \\
{$50\%$ } & {$70.77\%|74.71\%$} & {$86.89\%|95.21\%$} & {$63.34\%|61.31\%$} & {$83.96\%|90.23\%$} & {$70.08\%|70.12\%$} & {$92.58\%|97.54\%$} \\
{$60\%$ } & {$60.17\%|66.46\%$} & {$68.92\%|84.61\%$} & {$57.60\%|56.66\%$} & {$76.53\%|88.43\%$} & {$61.62\%|62.71\%$} & {$92.65\%|97.45\%$} \\
{$70\%$ } & {$52.46\%|55.14\%$} & {$34.31\%|75.71\%$} & {$56.39\%|52.79\%$} & {$73.90\%|89.45\%$} & {$53.58\%|52.91\%$} & {$92.38\%|97.48\%$} \\
{$80\%$ } & {$50.66\%|51.35\%$} & {$7.58\%|60.71\%$} & {$54.11\%|51.55\%$} & {$62.71\%|88.65\%$} & {$50.26\%|51.10\%$} & {$89.17\%|97.45\%$} \\
{$90\%$ } & {$50.66\%|50.36\%$} & {$7.58\%|51.00\%$} & {$52.00\%|50.47\%$} & {$58.77\%|89.54\%$} & {$49.67\%|49.07\%$} & {$76.90\%|82.74\%$} \\
\hline
\end{tabular}
\label{Table:pruning_NCI1}
\end{center}
\end{table*}

\begin{table*}[!htpb]

 \centering
 \caption{Watermarking performance on node classification task after model pruning (Cora).}
\begin{center}
\begin{tabular}{ccccccc}
 \hline
\multirow{2}{*}{Pruning rate} & \multicolumn{2}{c}{GCN} & \multicolumn{2}{c}{GAT} & \multicolumn{2}{c}{GraphSAGE} \\
\cline{2-7}
 & {Test Acc.} & {Watermark Acc.} & {Test Acc.} & {Watermark Acc.} & {Test Acc.} & {Watermark Acc.}\\
\hline
{$10\%$} & {$86.51\%$} & {$99.51\%$} & {$85.23\%$} & {$97.05\%$} & {$83.40\%$} & {$99.62\%$} \\
{$20\%$} & {$86.49\%$} & {$98.50\%$} & {$85.22\%$} & {$96.80\%$} & {$83.37\%$} & {$99.58\%$} \\
{$30\%$} & {$86.24\%$} & {$98.27\%$} & {$85.26\%$} & {$96.46\%$} & {$83.24\%$} & {$99.23\%$} \\
{$40\%$} & {$85.68\%$} & {$98.20\%$} & {$85.24\%$} & {$95.80\%$} & {$82.45\%$} & {$98.12\%$} \\
{$50\%$} & {$83.63\%$} & {$97.39\%$} & {$85.29\%$} & {$95.55\%$} & {$82.35\%$} & {$98.08\%$} \\
{$60\%$} & {$82.60\%$} & {$97.14\%$} & {$85.28\%$} & {$94.80\%$} & {$80.94\%$} & {$98.04\%$} \\
{$70\%$} & {$82.52\%$} & {$96.70\%$} & {$85.25\%$} & {$93.69\%$} & {$80.37\%$} & {$71.48\%$} \\
{$80\%$} & {$82.41\%$} & {$96.40\%$} & {$85.19\%$} & {$93.46\%$} & {$78.20\%$} & {$42.48\%$} \\
{$90\%$} & {$82.20\%$} & {$90.64\%$} & {$84.97\%$} & {$93.30\%$} & {$72.00\%$} & {$34.64\%$} \\
\hline
\end{tabular}
\label{Table:pruning_Cora}
\end{center}
\end{table*}

\textbf{Robustness against randomized subsampling.}
Randomized smoothing~\cite{cohen2019certified} is a state-of-the-art technique for building robust machine learning. For binary data, a randomized smoothing method based on randomized subsampling can achieve promising certified robustness~\cite{zhang2018protecting}. Here, we explore the robustness of our watermarking method against randomized subsampling~\cite{xi2021graph}.
In particular, we apply a subsampling function over a given graph $G$ to create a set of subsampled graphs $G_{s_1}, G_{s_2},\ldots, G_{s_n}$ by keeping some randomly subsampled nodes in $G$ and removing the remaining nodes. We then feed the subsampled graphs to the watermarked model and take a majority voting of the predictions over such graphs as $G$'s final prediction. 
In the randomized subsampling technique, there is an important parameter $\beta$ (subsampling ratio) that specifies the randomization degree. For example, if $\beta=0.2$, for the graph classification task, we randomly keep $20\%$ of $G$'s nodes and remove the rest of the nodes, and for the node classification task, we randomly keep the $20\%$ of the nodes' features in the graph and set the remaining features to $0$. Similar to~\cite{xi2021graph}, in this paper, the randomized subsampling is only used to work on graphs instead of training smoothed models (thus, we use it to make robust samples).

\begin{figure}[!ht]
\centering
\begin{subfigure}{0.48\textwidth}
  \centering
  \includegraphics[width=\textwidth]{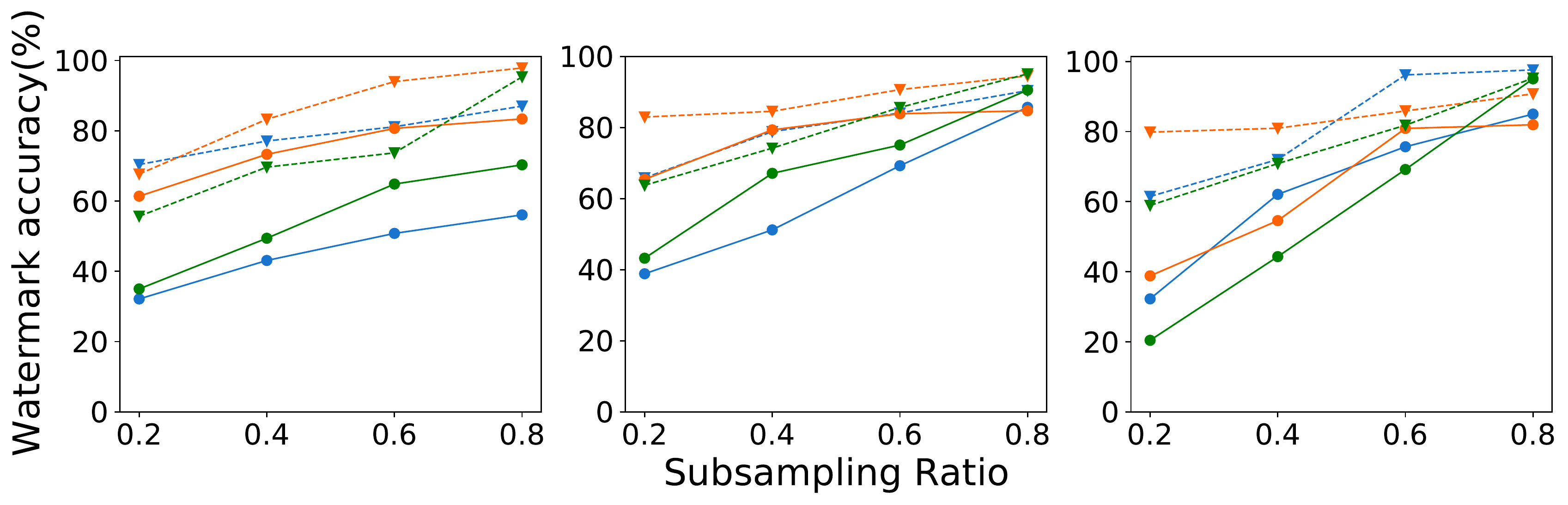}  
  \caption{watermark accuracy}
  \label{fig:randomly_preserve_graph_classification-a}
\end{subfigure}
\hfill
\begin{subfigure}{0.48\textwidth}
  \centering
  \includegraphics[width=\textwidth]{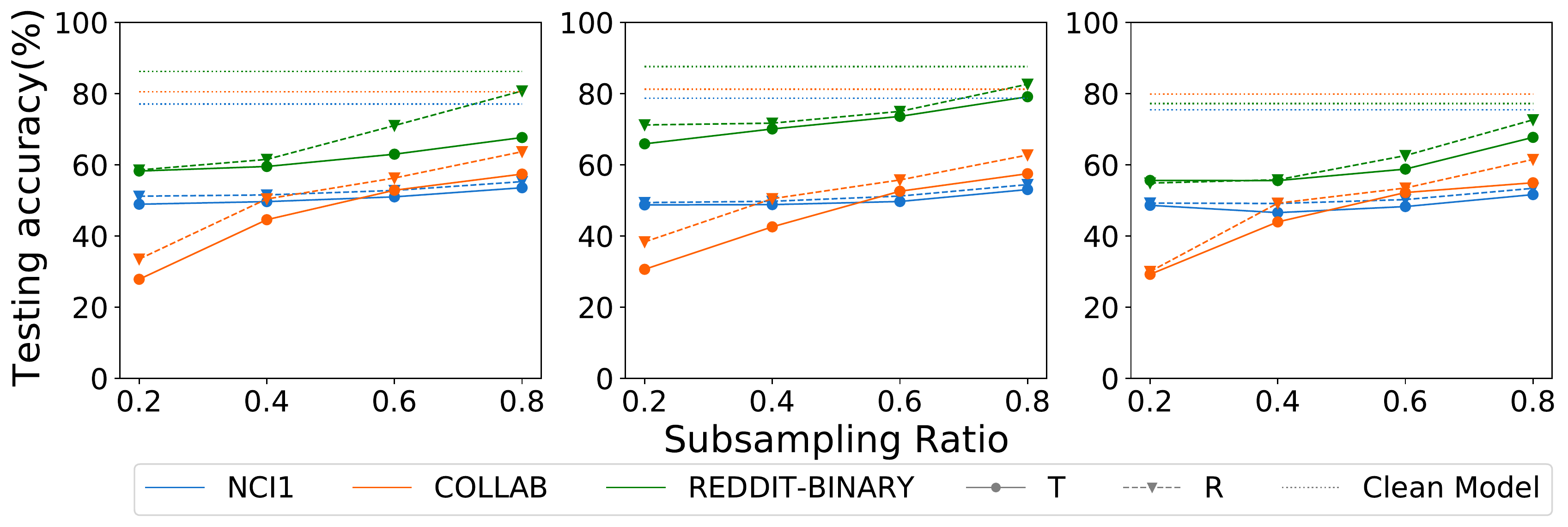}  
  \caption{testing accuracy}
  \label{fig:randomly_preserve_graph_classification-b}
\end{subfigure}
\caption{Watermarking performance for different subsampling ratios on graph classification task (DiffPool (left), GIN (center), GraphSAGE (right), T: first watermarked data generation strategy, R: second watermarked data generation strategy).}
\label{fig:randomly_preserve_graph_classification}
\end{figure}

\begin{figure}[htpb]
\centering
\begin{subfigure}{0.48\textwidth}
  \centering
  \includegraphics[width=\textwidth]{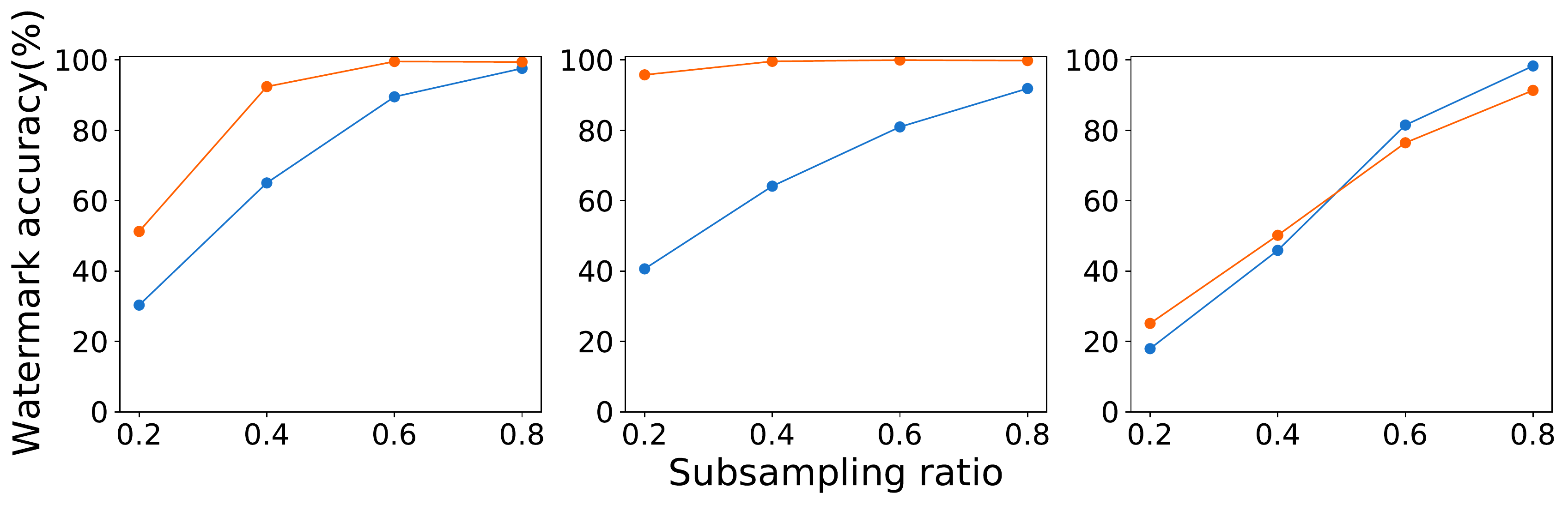}  
  \caption{watermark accuracy}
  \label{fig:randomly_preserve_node_classification-a}
\end{subfigure}
\hfill
\begin{subfigure}{0.48\textwidth}
  \centering
  \includegraphics[width=\textwidth]{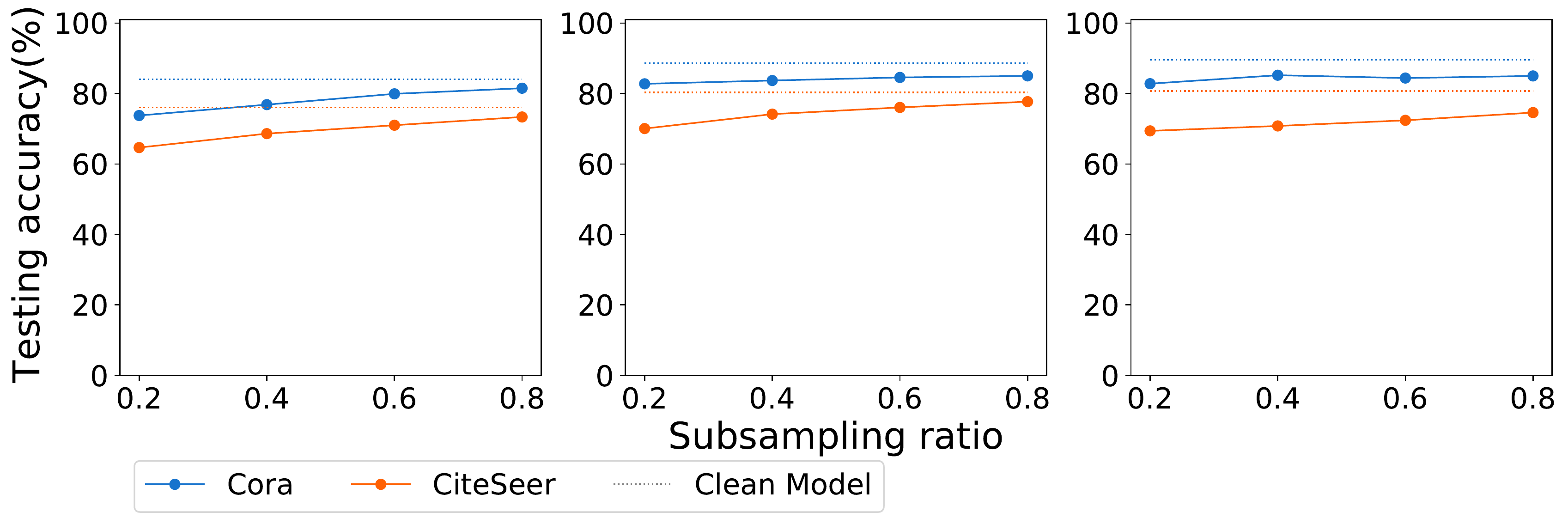}  
  \caption{testing accuracy}
  \label{fig:randomly_preserve_node_classification-b}
\end{subfigure}
\caption{Watermarking performance for different subsampling ratios on node classification task (GCN (left), GCN (center), GraphSAGE (right)).}
\label{fig:randomly_preserve_node_classification}
\end{figure}

Figures~\ref{fig:randomly_preserve_graph_classification} and~\ref{fig:randomly_preserve_node_classification} show the watermarking performance with different subsampling ratios in the graph classification task and the node classification task, respectively.  
We see that for the graph classification task, a decrease of $\beta$ decreases the watermark accuracy, and in most cases, the testing accuracy is significantly lower than the clean model's. 
However, for the node classification task, the reduction of the subsampling ratio leads to a significant drop in watermark accuracy but the testing accuracy is still close to the clean model. 
Therefore, randomized subsampling is not effective in attacking our watermarking mechanism for the graph classification task as the penalty in the testing accuracy is unacceptable. However, it is effective for the node classification task.

\textbf{Robustness against fine-pruning}
\begin{table*}[!htbp]
\centering
 \caption{Watermarking performance on graph classification task after fine-pruning (NCI1).}
\begin{center}
\begin{tabular}{ccccccc}
 \hline
\multirow{2}{*}{Pruning} & \multicolumn{2}{c}{DiffPool} & \multicolumn{2}{c}{GIN} & \multicolumn{2}{c}{GraphSAGE} \\
\cline{2-7}
  & {Test Acc. } & {Watermark Acc.} & {Test Acc.} & {Watermark Acc.} & {Test Acc.} & {Watermark Acc.}\\
 {rate} & ($D_{wm}^{t} | D_{wm}^{r}$) & ($D_{wm}^{t} | D_{wm}^{r}$) & ($D_{wm}^{t} | D_{wm}^{r}$) & ($D_{wm}^{t} | D_{wm}^{r}$) & ($D_{wm}^{t} | D_{wm}^{r}$) & ($D_{wm}^{t} | D_{wm}^{r}$) \\
\hline
{$10\%$} & {$81.32\%|77.20\%$} & {$95.69\%|95.44\%$} & {$82.32\%|79.18\%$} & {$90.24\%|90.61\%$} & {$74.96\%|75.71\%$} & {$93.46\%|97.92\%$} \\
{$20\%$} & {$81.00\%|77.43\%$} & {$94.46\%|92.10\%$} & {$82.01\%|79.12\%$} & {$90.19\%|89.45\%$} & {$75.00\%|75.51\%$} & {$93.39\%|97.29\%$} \\
{$30\%$} & {$81.16\%|77.52\%$} & {$93.46\%|88.64\%$} & {$81.57\%|78.99\%$} & {$90.07\%|89.46\%$} & {$74.58\%|75.50\%$} & {$92.79\%|97.62\%$} \\
{$40\%$} & {$80.35\%|77.43\%$} & {$93.38\%|88.44\%$} & {$81.12\%|79.06\%$} & {$90.05\%|89.54\%$} & {$73.74\%|75.43\%$} & {$92.90\%|97.02\%$} \\
{$50\%$} & {$79.20\%|77.58\%$} & {$89.78\%|87.61\%$} & {$80.10\%|78.98\%$} & {$90.14\%|89.32\%$} & {$72.37\%|75.66\%$} & {$92.71\%|96.80\%$} \\
{$60\%$} & {$76.83\%|77.48\%$} & {$88.36\%|86.80\%$} & {$78.93\%|78.92\%$} & {$90.03\%|89.17\%$} & {$68.01\%|75.17\%$} & {$93.10\%|96.75\%$} \\
{$70\%$} & {$75.14\%|77.48\%$} & {$88.10\%|83.30\%$} & {$75.74\%|78.70\%$} & {$89.89\%|89.10\%$} & {$61.86\%|74.25\%$} & {$92.66\%|96.82\%$} \\
{$80\%$} & {$69.73\%|77.20\%$} & {$81.67\%|82.87\%$} & {$72.01\%|77.78\%$} & {$89.93\%|89.30\%$} & {$55.62\%|73.95\%$} & {$93.05\%|96.77\%$} \\
{$90\%$} & {$66.71\%|75.21\%$} & {$67.42\%|82.45\%$} & {$66.93\%|75.89\%$} & {$82.64\%|83.06\%$} & {$50.71\%|66.37\%$} & {$92.91\%|73.55\%$} \\
\hline
\end{tabular}
\label{Table:fp_NCI1}
\end{center}
\end{table*}
\begin{table*}[!ht]
 \centering
 \caption{Watermarking performance on the node classification task after fine-pruning (GraphSAGE).}
\begin{center}
\begin{tabular}{ccccc}
 \hline
\multirow{2}{*}{Pruning rate} & \multicolumn{2}{c}{Cora} & \multicolumn{2}{c}{CiteSeer} \\
\cline{2-5}
 & {Test Acc.} & {Watermark Acc.} & {Test Acc.} & {Watermark Acc.} \\
\hline
{$10\%$} & {$84.04\%$} & {$99.70\%$} & {$72.12\%$} & {$99.27\%$} \\
{$20\%$} & {$83.45\%$} & {$99.83\%$} & {$71.37\%$} & {$99.49\%$} \\
{$30\%$} & {$83.06\%$} & {$99.53\%$} & {$71.00\%$} & {$99.14\%$} \\
{$40\%$} & {$83.45\%$} & {$99.23\%$} & {$71.56\%$} & {$99.59\%$} \\
{$50\%$} & {$83.26\%$} & {$99.07\%$} & {$71.93\%$} & {$99.65\%$} \\
{$60\%$} & {$82.48\%$} & {$98.06\%$} & {$71.56\%$} & {$88.96\%$} \\
{$70\%$} & {$83.06\%$} & {$69.18\%$} & {$71.19\%$} & {$51.60\%$} \\
{$80\%$} & {$71.70\%$} & {$36.24\%$} & {$61.93\%$} & {$16.75\%$} \\
{$90\%$} & {$64.87\%$} & {$26.71\%$} & {$56.55\%$} & {$8.86\%$} \\
\hline
\end{tabular}
\label{Table:fp_node}
\end{center}
\end{table*}
Fine-pruning is proposed as an effective defense against backdoor attacks on deep neural networks, combining two promising defenses, pruning and fine-tuning~\cite{liu2018fine}. 
As shown in the experimental results for the two defenses above, i.e., fine-tuning and pruning, neither is sufficient to eliminate the watermarking function. Therefore, here we assume the plagiarizer applies fine-pruning, a more effective defense, to train a new model on top of the stolen model. We follow the settings in the experiments of two defenses before, i.e., model pruning and fine-tuning. The pruning ratio is set from $10\%$ to $90\%$. 
We also take the results for NCI1 and Cora as examples, as shown in Tables~\ref{Table:fp_NCI1} and~\ref{Table:fp_node}, respectively. The results for other datasets are presented in Appendix~\ref{appendix:fine_pruning}. 
We can observe that the test accuracy of the second watermarked data generation strategy reduces slightly, which is different from the model pruning, where the test accuracy drops significantly. Thus, fine-pruning is a more powerful technique than model pruning for the plagiarizer to try to steal the model while keeping the model's performance on the original task. 
However, as we can also see from the results on the NCI1 dataset, with a pruning rate of $80\%$, the watermark accuracy decreases less than $2\%$ for the GIN and GraphSAGE models. For the DiffPool model, with the increasing of the pruning rate, the watermark accuracies for two watermarked data generation strategies drop to $67.42\%$ and $82.45\%$, respectively. However, they are still more than the corresponding thresholds, i.e., $53.5\%$ and $72.0\%$, respectively. 
As for the node classification task, based on the explanation in the fine-tuning experiments, we show the results for the GraphSAGE model. We can see from Table~\ref{Table:fp_node} that when half of the parameters are pruned in the fine-pruning, the drop on the watermark accuracy is less than $1\%$ for both datasets. With the pruning rate continuously increasing to $70\%$, the watermark accuracy decreases obviously, i.e., it drops to $69.18\%$ and $51.60\%$ for Cora and CiteSeer respectively, but it is still higher than the verification threshold. When the pruning rate is higher than $70\%$, the watermark accuracy reduces below the verification threshold but simultaneously, the test accuracy drops dramatically.  
Therefore, even if the plagiarizer performs fine-pruning to train a new model, our watermarking mechanism can still verify the ownership of the model.






\section{Conclusions and Future Work}
\label{sec:conclusions}

This paper proposed a watermarking framework for GNNs, which includes generating watermarked data with different strategies, embedding the watermark into the host model through training, and verifying the ownership of the suspicious model using previously generated watermarked data. We designed a watermarking mechanism for two GNN applications: the graph classification task and the node classification task, and provided statistical analysis for the model ownership verification results. 
We conducted a comprehensive evaluation of our watermarking framework on different datasets and models. We demonstrated that our method can achieve powerful watermarking performance while having a negligible effect on the host model's original task. 
We further show that our method is robust against a model extraction attack and four state-of-the-art defenses for backdoor attacks: randomized subsampling, fine-tuning, model pruning, and fine-pruning. 
For future work, we are interested in exploring methods that are more robust, e.g., against randomized smoothing for the node classification task. We will also study embedding watermarks into various types of GNNs besides node and graph-level tasks. 

\bibliographystyle{plain}
\bibliography{sample.bib}

\appendices

\begin{figure*}[!ht]
\centering
\includegraphics[width=0.8\textwidth, page=1]{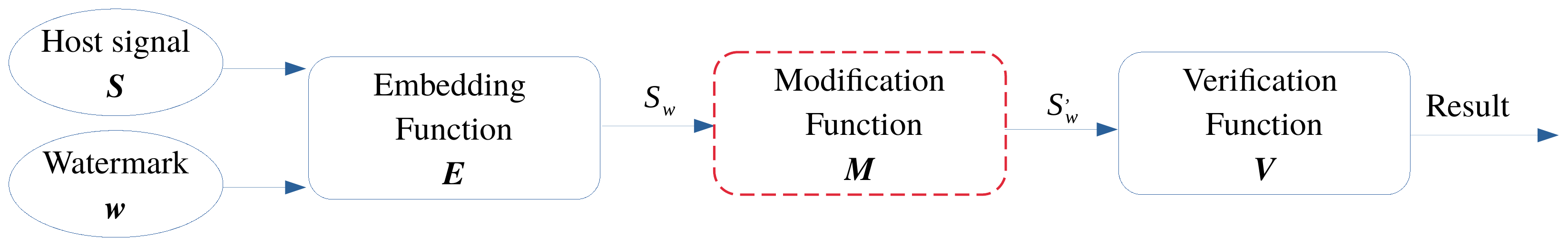}
\caption{Digital watermarking life cycle.}
\label{fig:watermark_life_cycle}
\end{figure*}

\section{Digital Watermarking Life Cycle}
\label{appendix:digital_watermarking_life_cycle}

Figure~\ref{fig:watermark_life_cycle} shows a typical digital watermarking life cycle. In the embedding step, an embedding algorithm $E$ embeds the watermark into the host signal to generate the watermarked data $S_w$. After embedding, the watermarked data is transferred or modified (dashed frame, this part is optional). During the watermark verification step, a verification algorithm is applied to attempt to extract the watermark from the watermarked signal. If the extracted watermark is equal to or within the acceptable distance to the original watermark, we can confirm that the signal is the protected signal.

\section{Statistic Guarantee of Ownership Verification}
\label{appendix:t_test}
The watermark accuracy of $10$ watermarked models and clean models for other datasets and models are shown in Tables~\ref{Table:t_test_g_t_appendix},~\ref{Table:t_test_g_r_appendix}, and~\ref{Table:t_test_n_appendix}. The corresponding $t$ statistics and $t$ critical values are also presented in these tables. 
As we can observe for each dataset and model, $t$ statistic is larger than the $t$ critical value, which supports the rejection of the null hypothesis $\mathcal{H}_0$ in~\ref{subsec:ownership_verification}. 

\begin{table*}[!ht]
 \caption{Accuracy of the watermarked models and clean models on $D_w^t$ for graph classification task ($n=10$).}
\begin{center}
\begin{tabular}{ccccccccccccccc}
 \hline
 {Setting} & {Models} & \multicolumn{10}{c}{Watermark Accuracy (\%)} & $t$ & $\nu$ & $t_{\tau}$ \\
 \hline
\multirow{2}{*}{NCI1\_GIN} & $m_c$ & $0.21$ & $3.17$ & $8.00$ & $13.78$ & $14.47$ & $8.27$ & $16.00$ & $9.02$ & $8.44$ & $12.69$ & \multirow{2}{*}{$26.04$} & \multirow{2}{*}{$14$} & \multirow{2}{*}{$2.145$} \\
\cline{2-12}
& $m_w$ & $94.42$ & $89.40$ & $85.22$ & $79.26$ & $83.98$ & $99.40$ & $99.73$ & $99.07$ & $85.91$ & $77.75$ & & & \\
\hline
\multirow{2}{*}{NCI1\_.SAGE} & $m_c$ & $0.10$ & $0.40$ & $1.64$ & $0.49$ & $3.65$ & $0.28$ & $0.87$ & $9.58$ & $0.39$ & $0.79$ & \multirow{2}{*}{$59.10$} & \multirow{2}{*}{$16$} & \multirow{2}{*}{$2.120$} \\
\cline{2-12}
& $m_w$ & $94.21$ & $93.16$ & $93.43$ & $87.50$ & $89.99$ & $99.98$ & $95.40$ & $87.47$ & $94.54$ & $90.02$ & & & \\
\hline
\multirow{2}{*}{COLLAB\_Diff.} & $m_c$ & $13.63$ & $10.84$ & $7.39$ & $7.62$ & $10.46$ & $12.47$ & $20.72$ & $11.49$ & $10.44$ & $7.06$ & \multirow{2}{*}{$39.17$} & \multirow{2}{*}{$17$} & \multirow{2}{*}{$2.110$} \\
\cline{2-12}
& $m_w$ & $76.19$ & $83.51$ & $85.16$ & $92.91$ & $82.83$ & $87.42$ & $84.53$ & $80.84$ & $84.51$ & $82.66$ & & & \\
\hline
\multirow{2}{*}{COLLAB\_GIN} & $m_c$ & $14.17$ & $9.88$ & $9.01$ & $16.89$ & $21.23$ & $17.65$ & $19.71$ & $14.37$ & $7.67$ & $16.82$ & \multirow{2}{*}{$25.41$} & \multirow{2}{*}{$15$} & \multirow{2}{*}{$2.131$} \\
\cline{2-12}
& $m_w$ & $84.80$ & $91.09$ & $77.28$ & $73.57$ & $88.93$ & $91.01$ & $91.73$ & $78.57$ & $76.35$ & $92.06$ & & & \\
\hline
\multirow{2}{*}{COLLAB\_.SAGE} & $m_c$ & $14.67$ & $17.66$ & $16.65$ & $17.59$ & $11.96$ & $22.51$ & $7.85$ & $17.56$ & $8.23$ & $13.41$ & \multirow{2}{*}{$37.06$} & \multirow{2}{*}{$17$} & \multirow{2}{*}{$2.110$} \\
\cline{2-12}
& $m_w$ & $85.60$ & $80.34$ & $83.54$ & $79.68$ & $82.83$ & $77.78$ & $81.52$ & $83.11$ & $88.24$ & $89.05$ & & & \\
\hline
\multirow{2}{*}{REDDIT.\_Diff.} & $m_c$ & $11.99$ & $11.08$ & $8.65$ & $7.33$ & $9.61$ & $0.87$ & $5.70$ & $3.96$ & $9.57$ & $12.35$ & \multirow{2}{*}{$49.70$} & \multirow{2}{*}{$18$} & \multirow{2}{*}{$2.101$} \\
\cline{2-12}
& $m_w$ & $83.48$ & $94.49$ & $94.76$ & $88.66$ & $89.85$ & $93.03$ & $88.23$ & $92.65$ & $86.49$ & $92.75$ & & & \\
\hline
\multirow{2}{*}{REDDIT.\_GIN} & $m_c$ & $13.64$ & $15.99$ & $10.24$ & $13.19$ & $11.27$ & $11.55$ & $4.79$ & $14.16$ & $11.98$ & $2.04$ & \multirow{2}{*}{$32.86$} & \multirow{2}{*}{$16$} & \multirow{2}{*}{$2.120$} \\
\cline{2-12}
& $m_w$ & $92.55$ & $86.07$ & $97.01$ & $91.46$ & $77.52$ & $83.99$ & $80.15$ & $86.00$ & $91.73$ & $89.23$ & & & \\
\hline
\multirow{2}{*}{REDDIT.\_.SAGE} & $m_c$ & $0.05$ & $0.04$ & $0.21$ & $2.52$ & $1.64$ & $1.18$ & $2.74$ & $0.79$ & $0.40$ & $0.52$ & \multirow{2}{*}{$164.02$} & \multirow{2}{*}{$15$} & \multirow{2}{*}{$2.131$} \\
\cline{2-12}
& $m_w$ & $96.21$ & $98.81$ & $99.09$ & $99.61$ & $99.53$ & $97.46$ & $97.57$ & $94.68$ & $98.14$ & $99.09$ & & & \\
\hline
\end{tabular}
\label{Table:t_test_g_t_appendix}
\end{center}
\end{table*}

\begin{table*}[!ht]
 \caption{Accuracy of the watermarked models and clean models on $D_w^r$ for graph classification task ($n=10$).}
\begin{center}
\begin{tabular}{ccccccccccccccc}
 \hline
 {Setting} & {Models} & \multicolumn{10}{c}{Watermark Accuracy (\%)} & $t$ & $\nu$ & $t_{\tau}$ \\
 \hline
\multirow{2}{*}{NCI1\_Diff.} & $m_c$ & $51.03$ & $47.51$ & $43.06$ & $51.30$ & $43.81$ & $48.01$ & $39.36$ & $52.08$ & $41.38$ & $42.92$ & \multirow{2}{*}{$26.37$} & \multirow{2}{*}{$17$} & \multirow{2}{*}{$2.110$} \\
\cline{2-12}
& $m_w$ & $95.62$ & $90.33$ & $93.50$ & $94.34$ & $96.71$ & $99.28$ & $97.42$ & $99.06$ & $95.44$ & $87.54$ & & & \\
\hline
\multirow{2}{*}{NCI1\_GIN} & $m_c$ & $35.00$ & $47.02$ & $55.15$ & $52.48$ & $52.65$ & $34.15$ & $50.74$ & $40.17$ & $51.01$ & $45.20$ & \multirow{2}{*}{$17.71$} & \multirow{2}{*}{$13$} & \multirow{2}{*}{$2.160$} \\
\cline{2-12}
& $m_w$ & $89.76$ & $96.34$ & $89.76$ & $95.44$ & $99.48$ & $88.53$ & $95.14$ & $91.87$ & $99.27$ & $97.11$ & & & \\
\hline
\multirow{2}{*}{NCI1\_.SAGE} & $m_c$ & $48.61$ & $50.85$ & $41.02$ & $55.53$ & $47.87$ & $51.62$ & $46.47$ & $44.07$ & $44.18$ & $50.62$ & \multirow{2}{*}{$30.15$} & \multirow{2}{*}{$15$} & \multirow{2}{*}{$2.131$} \\
\cline{2-12}
& $m_w$ & $95.47$ & $93.67$ & $97.20$ & $99.33$ & $99.60$ & $99.95$ & $97.82$ & $98.72$ & $99.49$ & $91.52$ & & & \\
\hline
\multirow{2}{*}{COLLAB\_Diff.} & $m_c$ & $26.90$ & $30.49$ & $30.46$ & $26.37$ & $28.61$ & $34.08$ & $25.72$ & $24.33$ & $27.63$ & $30.37$ & \multirow{2}{*}{$54.22$} & \multirow{2}{*}{$17$} & \multirow{2}{*}{$2.110$} \\
\cline{2-12}
& $m_w$ & $99.02$ & $98.69$ & $93.04$ & $99.16$ & $94.89$ & $93.74$ & $99.45$ & $99.92$ & $93.71$ & $96.67$ & & & \\
\hline
\multirow{2}{*}{COLLAB\_GIN} & $m_c$ & $34.52$ & $30.76$ & $20.49$ & $26.68$ & $33.93$ & $33.44$ & $30.72$ & $31.67$ & $30.99$ & $36.68$ & \multirow{2}{*}{$30.81$} & \multirow{2}{*}{$17$} & \multirow{2}{*}{$2.110$} \\
\cline{2-12}
& $m_w$ & $91.82$ & $88.00$ & $99.25$ & $88.16$ & $89.55$ & $99.52$ & $92.86$ & $93.51$ & $99.26$ & $92.53$ & & & \\
\hline
\multirow{2}{*}{COLLAB\_.SAGE} & $m_c$ & $23.07$ & $29.23$ & $24.54$ & $24.79$ & $28.36$ & $32.10$ & $34.99$ & $28.20$ & $31.53$ & $25.01$ & \multirow{2}{*}{$49.00$} & \multirow{2}{*}{$14$} & \multirow{2}{*}{$2.145$} \\
\cline{2-12}
& $m_w$ & $99.59$ & $95.28$ & $93.24$ & $99.25$ & $98.10$ & $96.07$ & $99.98$ & $99.37$ & $99.67$ & $97.15$ & & & \\
\hline
\multirow{2}{*}{REDDIT.\_Diff.} & $m_c$ & $42.87$ & $41.59$ & $46.01$ & $42.85$ & $46.20$ & $40.40$ & $39.23$ & $41.22$ & $40.50$ & $47.81$ & \multirow{2}{*}{$44.89$} & \multirow{2}{*}{$17$} & \multirow{2}{*}{$2.110$} \\
\cline{2-12}
& $m_w$ & $95.58$ & $95.79$ & $99.27$ & $94.36$ & $97.29$ & $92.61$ & $99.16$ & $99.74$ & $99.95$ & $96.39$ & & & \\
\hline
\multirow{2}{*}{REDDIT.\_GIN} & $m_c$ & $49.90$ & $43.29$ & $46.39$ & $47.79$ & $45.60$ & $44.44$ & $48.16$ & $40.27$ & $46.86$ & $45.63$ & \multirow{2}{*}{$31.24$} & \multirow{2}{*}{$15$} & \multirow{2}{*}{$2.131$} \\
\cline{2-12}
& $m_w$ & $86.38$ & $92.03$ & $97.16$ & $99.92$ & $98.44$ & $99.27$ & $99.39$ & $95.35$ & $92.99$ & $97.14$ & & & \\
\hline
\multirow{2}{*}{REDDIT.\_.SAGE} & $m_c$ & $50.89$ & $43.93$ & $47.71$ & $47.44$ & $47.93$ & $45.83$ & $53.12$ & $53.68$ & $44.45$ & $44.65$ & \multirow{2}{*}{$39.52$} & \multirow{2}{*}{$13$} & \multirow{2}{*}{$2.160$} \\
\cline{2-12}
& $m_w$ & $93.85$ & $99.33$ & $97.34$ & $96.49$ & $97.49$ & $96.15$ & $97.42$ & $99.39$ & $99.17$ & $99.47$ & & & \\
\hline
\end{tabular}
\label{Table:t_test_g_r_appendix}
\end{center}
\end{table*}

\begin{table*}[!ht]
 \caption{Accuracy of the watermarked models and clean models on watermarked data for node classification task ($n=10$).}
\begin{center}
\begin{tabular}{ccccccccccccccc}
 \hline
 {Setting} & {Models} & \multicolumn{10}{c}{Watermark Accuracy (\%)} & $t$ & $\nu$ & $t_{\tau}$ \\
 \hline
\multirow{2}{*}{Cora\_GCN} & $m_c$ & $0.33$ & $0.17$ & $6.25$ & $3.23$ & $6.17$ & $3.83$ & $4.03$ & $0.20$ & $0.93$ & $2.96$ & \multirow{2}{*}{$88.78$} & \multirow{2}{*}{$17$} & \multirow{2}{*}{$2.110$} \\
\cline{2-12}
& $m_w$ & $99.66$ & $93.14$ & $99.42$ & $94.37$ & $96.97$ & $99.91$ & $98.11$ & $95.73$ & $98.47$ & $99.86$ & & & \\
\hline
\multirow{2}{*}{Cora\_GAT} & $m_c$ & $10.39$ & $3.73$ & $10.14$ & $0.87$ & $5.81$ & $10.66$ & $4.66$ & $2.63$ & $9.74$ & $7.61$ & \multirow{2}{*}{$53.51$} & \multirow{2}{*}{$17$} & \multirow{2}{*}{$2.110$} \\
\cline{2-12}
& $m_w$ & $92.30$ & $99.90$ & $99.73$ & $87.90$ & $97.98$ & $95.01$ & $99.24$ & $95.55$ & $99.57$ & $95.16$ & & & \\
\hline
\multirow{2}{*}{Cora\_.SAGE} & $m_c$ & $1.15$ & $4.94$ & $1.61$ & $4.57$ & $3.12$ & $4.81$ & $0.67$ & $0.22$ & $0.52$ & $1.25$ & \multirow{2}{*}{$89.19$} & \multirow{2}{*}{$15$} & \multirow{2}{*}{$2.131$} \\
\cline{2-12}
& $m_w$ & $97.52$ & $93.24$ & $92.44$ & $99.97$ & $95.22$ & $99.94$ & $99.17$ & $96.25$ & $97.99$ & $99.80$ & & & \\
\hline
\multirow{2}{*}{Cite.\_GCN} & $m_c$ & $0.92$ & $2.06$ & $3.22$ & $10.08$ & $0.72$ & $1.26$ & $4.11$ & $6.76$ & $7.70$ & $0.23$ & \multirow{2}{*}{$78.00$} & \multirow{2}{*}{$13$} & \multirow{2}{*}{$2.160$} \\
\cline{2-12}
& $m_w$ & $99.96$ & $99.86$ & $95.94$ & $95.17$ & $99.07$ & $97.59$ & $96.30$ & $99.71$ & $97.81$ & $99.09$ & & & \\
\hline
\multirow{2}{*}{Cite.\_GAT} & $m_c$ & $2.50$ & $0.55$ & $0.86$ & $3.32$ & $0.61$ & $2.03$ & $1.41$ & $2.56$ & $0.37$ & $0.37$ & \multirow{2}{*}{$129.83$} & \multirow{2}{*}{$13$} & \multirow{2}{*}{$2.160$} \\
\cline{2-12}
& $m_w$ & $99.30$ & $99.13$ & $99.01$ & $97.78$ & $98.20$ & $94.13$ & $95.87$ & $94.69$ & $99.99$ & $99.20$ & & & \\
\hline
\multirow{2}{*}{Cite.\_.SAGE} & $m_c$ & $0.99$ & $1.05$ & $0.73$ & $1.14$ & $0.81$ & $0.35$ & $0.34$ & $0.28$ & $1.33$ & $0.22$ & \multirow{2}{*}{$381.65$} & \multirow{2}{*}{$14$} & \multirow{2}{*}{$2.145$} \\
\cline{2-12}
& $m_w$ & $99.75$ & $97.83$ & $99.95$ & $99.41$ & $99.74$ & $99.85$ & $98.22$ & $99.13$ & $99.36$ & $99.01$ & & & \\
\hline
\end{tabular}
\label{Table:t_test_n_appendix}
\end{center}
\end{table*}

\section{Additional Experimental Results}
\label{appendix:additional_exps}

\subsection{Model Pruning}
\label{appendix:model_pruning}

The watermarking performance after model pruning on the COLLAB, REDDIT-BINARY, and CiteSeer datasets is shown in Tables~\ref{Table:pruning_COLLAB},~\ref{Table:pruning_REDDIT_BINARY}, and~\ref{Table:pruning_CiteSeer}, respectively. For the COLLAB dataset, when the pruning rate is less than $50\%$, the watermark accuracy drops slightly. With a pruning rate of more than $50\%$, the watermarking accuracy decreases significantly as well as the testing accuracy for all models. The results for REDDIT-BINARY have the same phenomenon.
As for the CiteSeer dataset, even with $90\%$ of the parameters pruned, the watermark accuracy on the GCN and GAT models is still high, i.e., more than $99\%$. On the other hand, with more than $50\%$ of the parameters pruned, the watermark accuracy on the GraphSAGE model drops dramatically, but the testing accuracy decreases significantly as well.
Therefore, these results further verify that our watermarking mechanism is robust to model pruning, but the plagiarizer can still eliminate our watermarks with the cost of high accuracy drop in the main task.

\begin{table*}[!ht]
 \centering
 \caption{Watermarking performance on graph classification task after model pruning (COLLAB).}
\begin{center}
\begin{tabular}{ccccccc}
 \hline
\multirow{2}{*}{Pruning} & \multicolumn{2}{c}{DiffPool} & \multicolumn{2}{c}{GIN} & \multicolumn{2}{c}{GraphSAGE} \\
\cline{2-7}
  & {Test Acc. } & {Watermark Acc.} & {Test Acc.} & {Watermark Acc.} & {Test Acc.} & {Watermark Acc.}\\
 {rate} & ($D_{wm}^{t} | D_{wm}^{r}$) & ($D_{wm}^{t} | D_{wm}^{r}$) & ($D_{wm}^{t} | D_{wm}^{r}$) & ($D_{wm}^{t} | D_{wm}^{r}$) & ($D_{wm}^{t} | D_{wm}^{r}$) & ($D_{wm}^{t} | D_{wm}^{r}$) \\
\hline
{$10\%$ } & {$80.36\%|80.69\%$} & {$85.30\%|98.01\%$} & {$81.39\%|81.66\%$} & {$85.24\%|94.08\%$} & {$79.65\%|79.89\%$} & {$82.89\%|97.19\%$} \\
{$20\%$ } & {$80.12\%|80.55\%$} & {$85.35\%|97.88\%$} & {$80.71\%|81.49\%$} & {$84.70\%|94.02\%$} & {$79.59\%|79.90\%$} & {$82.91\%|97.10\%$} \\
{$30\%$ } & {$79.89\%|80.15\%$} & {$84.71\%|97.91\%$} & {$79.28\%|81.44\%$} & {$83.49\%|93.56\%$} & {$79.37\%|79.69\%$} & {$82.90\%|96.39\%$} \\
{$40\%$ } & {$78.43\%|79.28\%$} & {$84.78\%|97.53\%$} & {$77.13\%|80.84\%$} & {$83.34\%|92.46\%$} & {$78.96\%|79.57\%$} & {$82.72\%|96.50\%$} \\
{$50\%$ } & {$70.29\%|76.36\%$} & {$70.79\%|84.06\%$} & {$73.46\%|79.24\%$} & {$78.70\%|69.19\%$} & {$76.97\%|79.34\%$} & {$81.45\%|96.48\%$} \\
{$60\%$ } & {$59.58\%|64.80\%$} & {$64.05\%|63.23\%$} & {$71.12\%|76.55\%$} & {$78.16\%|43.72\%$} & {$72.91\%|77.91\%$} & {$76.84\%|96.43\%$} \\
{$70\%$ } & {$50.64\%|47.63\%$} & {$61.47\%|35.44\%$} & {$67.22\%|70.30\%$} & {$75.53\%|43.72\%$} & {$62.64\%|72.64\%$} & {$62.68\%|96.32\%$} \\
{$80\%$ } & {$47.93\%|36.60\%$} & {$63.13\%|35.44\%$} & {$60.99\%|63.52\%$} & {$73.98\%|43.72\%$} & {$47.54\%|57.77\%$} & {$44.52\%|82.91\%$} \\
{$90\%$ } & {$47.90\%|32.53\%$} & {$63.09\%|28.50\%$} & {$55.86\%|42.34\%$} & {$74.33\%|30.99\%$} & {$35.17\%|37.26\%$} & {$36.40\%|62.39\%$} \\
\hline
\end{tabular}
\label{Table:pruning_COLLAB}
\end{center}
\end{table*}

\begin{table*}[!ht]
 \centering
 \caption{Watermarking performance on graph classification task after model pruning (REDDIT-BINARY).}
\begin{center}
\begin{tabular}{ccccccc}
 \hline
\multirow{2}{*}{Pruning} & \multicolumn{2}{c}{DiffPool} & \multicolumn{2}{c}{GIN} & \multicolumn{2}{c}{GraphSAGE} \\
\cline{2-7}
  & {Test Acc. } & {Watermark Acc.} & {Test Acc.} & {Watermark Acc.} & {Test Acc.} & {Watermark Acc.}\\
 {rate} & ($D_{wm}^{t} | D_{wm}^{r}$) & ($D_{wm}^{t} | D_{wm}^{r}$) & ($D_{wm}^{t} | D_{wm}^{r}$) & ($D_{wm}^{t} | D_{wm}^{r}$) & ($D_{wm}^{t} | D_{wm}^{r}$) & ($D_{wm}^{t} | D_{wm}^{r}$) \\
\hline
{$10\%$} & {$86.75\%|86.55\%$} & {$90.05\%|98.40\%$} & {$87.41\%|87.05\%$} & {$91.81\%|99.09\%$} & {$78.30\%|77.94\%$} & {$99.35\%|98.87\%$} \\
{$20\%$} & {$86.78\%|86.58\%$} & {$90.15\%|97.99\%$} & {$87.20\%|87.11\%$} & {$91.34\%|98.84\%$} & {$78.33\%|77.89\%$} & {$99.30\%|99.09\%$} \\
{$30\%$} & {$86.79\%|86.24\%$} & {$89.98\%|98.40\%$} & {$87.30\%|86.84\%$} & {$91.14\%|98.69\%$} & {$78.33\%|77.77\%$} & {$99.48\%|99.00\%$} \\
{$40\%$} & {$86.58\%|86.42\%$} & {$89.79\%|98.41\%$} & {$87.26\%|85.92\%$} & {$91.24\%|98.78\%$} & {$77.96\%|77.74\%$} & {$99.00\%|98.82\%$} \\
{$50\%$} & {$86.91\%|86.55\%$} & {$89.10\%|98.63\%$} & {$84.76\%|84.98\%$} & {$91.75\%|98.77\%$} & {$77.70\%|77.48\%$} & {$99.05\%|98.08\%$} \\
{$60\%$} & {$86.43\%|85.70\%$} & {$89.43\%|97.73\%$} & {$81.45\%|83.57\%$} & {$88.26\%|96.09\%$} & {$76.85\%|77.56\%$} & {$99.66\%|98.91\%$} \\
{$70\%$} & {$82.61\%|82.89\%$} & {$88.94\%|93.02\%$} & {$77.99\%|80.72\%$} & {$77.11\%|76.73\%$} & {$73.40\%|76.34\%$} & {$99.43\%|99.20\%$} \\
{$80\%$} & {$72.62\%|74.39\%$} & {$80.58\%|87.82\%$} & {$73.48\%|76.22\%$} & {$70.71\%|82.72\%$} & {$64.95\%|72.43\%$} & {$99.48\%|98.81\%$} \\
{$90\%$} & {$60.69\%|63.94\%$} & {$31.76\%|66.76\%$} & {$68.11\%|70.56\%$} & {$67.31\%|77.22\%$} & {$53.79\%|59.67\%$} & {$98.36\%|99.08\%$} \\
\hline
\end{tabular}
\label{Table:pruning_REDDIT_BINARY}
\end{center}
\end{table*}

\begin{table*}[!ht]
 \centering
 \caption{Watermarking performance on node classification task after model pruning (CiteSeer)}
\begin{center}
\begin{tabular}{ccccccc}
 \hline
\multirow{2}{*}{Pruning rate} & \multicolumn{2}{c}{GCN} & \multicolumn{2}{c}{GAT} & \multicolumn{2}{c}{GraphSAGE} \\
\cline{2-7}
 & {Test Acc.} & {Watermark Acc.} & {Test Acc.} & {Watermark Acc.} & {Test Acc.} & {Watermark Acc.}\\
\hline
{$10\%$} & {$74.50\%$} & {$99.45\%$} & {$78.55\%$} & {$99.59\%$} & {$80.25\%$} & {$93.90\%$} \\
{$20\%$} & {$74.42\%$} & {$99.42\%$} & {$78.42\%$} & {$99.55\%$} & {$80.34\%$} & {$93.82\%$} \\
{$30\%$} & {$74.47\%$} & {$99.40\%$} & {$78.57\%$} & {$99.57\%$} & {$80.24\%$} & {$93.86\%$} \\
{$40\%$} & {$74.45\%$} & {$99.35\%$} & {$78.49\%$} & {$99.60\%$} & {$80.28\%$} & {$93.90\%$} \\
{$50\%$} & {$74.41\%$} & {$99.31\%$} & {$78.51\%$} & {$99.52\%$} & {$80.45\%$} & {$93.80\%$} \\
{$60\%$} & {$74.35\%$} & {$99.24\%$} & {$78.35\%$} & {$99.59\%$} & {$75.32\%$} & {$82.79\%$} \\
{$70\%$} & {$74.22\%$} & {$99.01\%$} & {$78.32\%$} & {$99.53\%$} & {$74.27\%$} & {$49.52\%$} \\
{$80\%$} & {$73.93\%$} & {$99.74\%$} & {$78.08\%$} & {$99.43\%$} & {$73.20\%$} & {$16.02\%$} \\
{$90\%$} & {$72.93\%$} & {$99.95\%$} & {$77.59\%$} & {$99.87\%$} & {$72.00\%$} & {$8.46\%$} \\
\hline
\end{tabular}
\label{Table:pruning_CiteSeer}
\end{center}
\end{table*}

\subsection{Fine-pruning}
\label{appendix:fine_pruning}

The watermarking performance after fine-pruning on the COLLAB and REDDIT-BINARY datasets is shown in Tables~\ref{Table:fp_COLLAB} and~\ref{Table:fp_REDDIT_BINARY}, respectively. For the COLLAB dataset, with the increase in the pruning rate, the watermark accuracy gradually decreases. 
However, even with the $90\%$ pruning rate, most of the watermark accuracy is still higher than the threshold in Table~\ref{watermark_acc_threshold_graph}. 
The results for REDDIT-BINARY follow the same phenomenon.
Therefore, these results further verify that our watermarking mechanism is robust to fine-pruning.

\begin{table*}[!ht]
 \centering
 \caption{Watermarking performance on graph classification task after fine-pruning (COLLAB).}
\begin{center}
\begin{tabular}{ccccccc}
 \hline
\multirow{2}{*}{Pruning} & \multicolumn{2}{c}{DiffPool} & \multicolumn{2}{c}{GIN} & \multicolumn{2}{c}{GraphSAGE} \\
\cline{2-7}
  & {Test Acc. } & {Watermark Acc.} & {Test Acc.} & {Watermark Acc.} & {Test Acc.} & {Watermark Acc.}\\
 {rate} & ($D_{wm}^{t} | D_{wm}^{r}$) & ($D_{wm}^{t} | D_{wm}^{r}$) & ($D_{wm}^{t} | D_{wm}^{r}$) & ($D_{wm}^{t} | D_{wm}^{r}$) & ($D_{wm}^{t} | D_{wm}^{r}$) & ($D_{wm}^{t} | D_{wm}^{r}$) \\
\hline
{$10\%$} & {$71.77\%|92.62\%$} & {$85.37\%|97.95\%$} & {$69.55\%|87.42\%$} & {$85.21\%|94.66\%$} & {$69.07\%|92.06\%$} & {$82.93\%|96.59\%$} \\
{$20\%$} & {$72.07\%|92.71\%$} & {$85.37\%|80.59\%$} & {$69.52\%|87.56\%$} & {$85.03\%|88.29\%$} & {$69.39\%|91.98\%$} & {$82.93\%|95.19\%$} \\
{$30\%$} & {$70.50\%|93.07\%$} & {$85.33\%|71.91\%$} & {$68.04\%|87.48\%$} & {$83.97\%|87.96\%$} & {$69.20\%|92.00\%$} & {$82.91\%|85.19\%$} \\
{$40\%$} & {$65.29\%|93.08\%$} & {$85.33\%|71.88\%$} & {$65.80\%|87.07\%$} & {$81.24\%|87.77\%$} & {$69.38\%|91.93\%$} & {$82.91\%|87.12\%$} \\
{$50\%$} & {$60.62\%|93.28\%$} & {$85.31\%|71.74\%$} & {$62.80\%|86.84\%$} & {$81.08\%|56.46\%$} & {$69.20\%|91.84\%$} & {$82.90\%|86.02\%$} \\
{$60\%$} & {$56.44\%|93.25\%$} & {$85.22\%|71.66\%$} & {$57.29\%|86.10\%$} & {$80.45\%|50.09\%$} & {$65.23\%|91.87\%$} & {$82.15\%|85.77\%$} \\
{$70\%$} & {$53.53\%|93.35\%$} & {$84.73\%|63.22\%$} & {$53.28\%|86.47\%$} & {$79.86\%|49.53\%$} & {$60.01\%|91.59\%$} & {$82.00\%|85.02\%$} \\
{$80\%$} & {$52.39\%|93.16\%$} & {$84.46\%|63.14\%$} & {$52.93\%|85.95\%$} & {$78.57\%|43.72\%$} & {$54.77\%|91.00\%$} & {$81.41\%|83.92\%$} \\
{$90\%$} & {$52.45\%|92.70\%$} & {$84.20\%|63.09\%$} & {$52.44\%|84.64\%$} & {$78.04\%|43.38\%$} & {$51.94\%|89.87\%$} & {$80.23\%|73.79\%$} \\
\hline
\end{tabular}
\label{Table:fp_COLLAB}
\end{center}
\end{table*}

\begin{table*}[!ht]
 \centering
 \caption{Watermarking performance on graph classification task after fine-pruning (REDDIT-BINARY).}
\begin{center}
\begin{tabular}{ccccccc}
 \hline
\multirow{2}{*}{Pruning} & \multicolumn{2}{c}{DiffPool} & \multicolumn{2}{c}{GIN} & \multicolumn{2}{c}{GraphSAGE} \\
\cline{2-7}
  & {Test Acc. } & {Watermark Acc.} & {Test Acc.} & {Watermark Acc.} & {Test Acc.} & {Watermark Acc.}\\
 {rate} & ($D_{wm}^{t} | D_{wm}^{r}$) & ($D_{wm}^{t} | D_{wm}^{r}$) & ($D_{wm}^{t} | D_{wm}^{r}$) & ($D_{wm}^{t} | D_{wm}^{r}$) & ($D_{wm}^{t} | D_{wm}^{r}$) & ($D_{wm}^{t} | D_{wm}^{r}$) \\
\hline
{$10\%$} & {$83.29\%|95.52\%$} & {$90.20\%|98.59\%$} & {$84.83\%|93.35\%$} & {$92.14\%|99.04\%$} & {$73.76\%|88.37\%$} & {$99.51\%|99.21\%$} \\
{$20\%$} & {$83.02\%|95.52\%$} & {$88.02\%|96.13\%$} & {$84.47\%|93.03\%$} & {$90.73\%|96.31\%$} & {$73.67\%|87.86\%$} & {$99.96\%|98.90\%$} \\
{$30\%$} & {$83.20\%|95.52\%$} & {$87.54\%|96.08\%$} & {$85.97\%|92.82\%$} & {$86.78\%|98.35\%$} & {$73.85\%|87.77\%$} & {$99.28\%|99.32\%$} \\
{$40\%$} & {$84.84\%|95.67\%$} & {$86.45\%|95.65\%$} & {$85.10\%|93.06\%$} & {$86.67\%|97.62\%$} & {$73.94\%|87.83\%$} & {$99.80\%|99.20\%$} \\
{$50\%$} & {$84.45\%|95.91\%$} & {$86.11\%|95.60\%$} & {$84.65\%|93.09\%$} & {$87.13\%|97.19\%$} & {$73.88\%|88.07\%$} & {$99.68\%|98.54\%$} \\
{$60\%$} & {$84.72\%|96.15\%$} & {$74.72\%|94.74\%$} & {$83.51\%|93.20\%$} & {$66.76\%|82.40\%$} & {$73.49\%|87.86\%$} & {$99.46\%|98.80\%$} \\
{$70\%$} & {$83.62\%|95.94\%$} & {$74.04\%|93.58\%$} & {$83.00\%|93.65\%$} & {$65.47\%|82.07\%$} & {$72.60\%|87.56\%$} & {$99.40\%|98.42\%$} \\
{$80\%$} & {$82.66\%|95.73\%$} & {$71.52\%|92.76\%$} & {$80.18\%|93.71\%$} & {$35.34\%|81.18\%$} & {$72.30\%|88.58\%$} & {$99.50\%|98.33\%$} \\
{$90\%$} & {$78.17\%|95.94\%$} & {$54.68\%|84.63\%$} & {$73.74\%|93.74\%$} & {$28.87\%|79.52\%$} & {$69.07\%|90.00\%$} & {$99.52\%|98.39\%$} \\
\hline
\end{tabular}
\label{Table:fp_REDDIT_BINARY}
\end{center}
\end{table*}

\end{document}